\documentclass[10pt,twocolumn,letterpaper]{article}
\usepackage[pagenumbers]{cvpr}
\usepackage[T1]{fontenc}
\usepackage[utf8]{inputenc}
\usepackage{array}
\usepackage{tabularx}
\usepackage{placeins}
\usepackage{mathtools}

\usepackage{pgfplots}
\pgfplotsset{compat=1.18}
\definecolor{cvprblue}{rgb}{0.21,0.49,0.74}
\usepackage[pagebackref,breaklinks,colorlinks,allcolors=cvprblue]{hyperref}

\title{Pick, Don't Push: Emotion-Preserving Face Augmentation via Pre-Generation Noise Selection}
\author{%
Haonan Zhang$^{1}$\thanks{Equal contribution. Emails: \texttt{zhn19375264@buaa.edu.cn}, \texttt{yan\_hao@buaa.edu.cn}.}\quad
Hao Yan$^{1}$\footnotemark[1]\quad
Yue Xia$^{2}$\thanks{Email: \texttt{xiayy@stu.xjtu.edu.cn}.}\quad
Bowei Wang$^{1}$\thanks{Corresponding authors. Emails: \texttt{wangboy0716@163.com}, \texttt{tangwenzhong@buaa.edu.cn}.}\quad
Wenzhong Tang$^{1}$\footnotemark[3]\\[3pt]
$^{1}$Beihang University\qquad $^{2}$Xi'an Jiaotong University}

\hypersetup{
pdftitle={Pick, Don't Push: Emotion-Preserving Face Augmentation via Pre-Generation Noise Selection},
pdfauthor={Haonan Zhang, Hao Yan, Yue Xia, Bowei Wang, Wenzhong Tang},
pdfsubject={Research preprint}}

\begin{document}
\maketitle

\begin{abstract}
Rare emotions and subtle expression variations are difficult to collect at scale. Emotion-preserving face augmentation generates multiple emotion-consistent, high-quality, and visually distinct faces from one source image. We find that a frozen conditional generator already produces candidates highly similar to the source emotion representation, and that the effectiveness of a noise vector depends mainly on its interaction with the source. These interactions concentrate in a few directions and are predictable from source representations. We propose Source-Aware Face Augmentation (SAFA), which supervises a noise selector with a complete response matrix to determine source-specific noise sets before generation. Under cross-encoder evaluation, SAFA matches the representation similarity of exhaustive selection with about 6\% of its generations and yields more usable emotion-preserving samples under the same online generation budget. Its samples also improve facial expression recognition with scarce labels by 16.8\% in relative accuracy.
\end{abstract}

\begin{figure}[t]
    \centering
    \includegraphics[width=\linewidth]{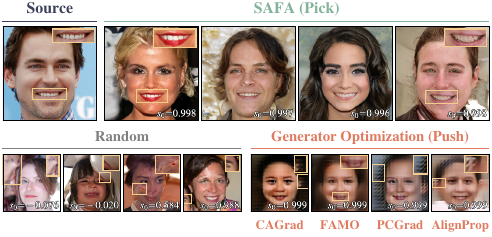}
    \caption{Generator optimization increases similarity under the training encoder, while noise selection produces faces that preserve the source emotion and pass quality assessment. Each of the three approaches generates four images from the same source. Random noise largely fails to preserve subtle expressions; optimizing the generator against $E_0$ increases representation similarity but produces structural artifacts; and faces generated from the noise vectors selected by SAFA before generation preserve the source emotion while producing distinct appearances. The representation similarity $s_0$ is shown at the bottom right of each image.}
    \label{fig:teaser}
\end{figure}

\section{Introduction}
\label{sec:intro}

Fine-grained facial expression recognition requires training data that cover subtle emotional differences. Two faces labeled ``happy'' may show slightly raised mouth corners or a broad smile, yet such variations are unevenly distributed in real data, and rare and compound emotions are scarce and costly to annotate. Emotion-preserving augmentation addresses this gap by generating new faces that share the emotion of a source image but differ in appearance. Since facial emotion can be captured by the low-dimensional representation of an expression encoder, the goal is to inject the source emotion representation into new faces. Most face generators, however, are conditioned on text or class labels and struggle to accept and preserve such a specific representation.

A direct way to preserve a source representation in generated images is to train the generator. Multi-objective optimization~\cite{ref1,ref2,ref3} and reward fine-tuning~\cite{ref4} introduce a representation similarity loss alongside the generation loss. In our experiments, these methods bring similarity under the training encoder close to saturation and also increase similarity under held-out encoders, but sharply reduce the number of samples that pass quality assessment. Moreover, this loss must backpropagate through both generation and encoding, making training computationally expensive.

In contrast, samples drawn from random noise retain the generator's original image quality without additional training. A single random sample typically has low similarity, but among the candidates generated for one source from a set of noise vectors, the top few far exceed the average. We therefore replace generator optimization with noise selection (pick, don't push); the key is to identify these noise vectors before generation.

We find that about 85\% of the variation in representation similarity across noise vectors for a fixed source arises from source-noise interaction effects, i.e., the same noise vector affects different sources differently. These effects concentrate in a few directions, and similar sources prefer similar noise vectors, so they are predictable from source representations. Based on this finding, we propose Source-Aware Face Augmentation (SAFA). SAFA conditions the generator with its native training objective and then freezes it. It records the representation similarities between training sources and random noise vectors as a response matrix, which supervises a lightweight noise selector. SAFA is thus a form of self-distillation that leaves the generator weights unchanged.

This design yields an unexpected result: prediction before generation achieves cross-encoder similarity comparable to measuring each image after generation, despite lower training similarity. Selection by per-image scores favors images that the encoder overestimates, whereas SAFA learns from low-rank structure shared across sources. Unlike Good Seed~\cite{ref5} and NoiseQuery~\cite{ref6}, which treat the effectiveness of noise as a property of the noise itself, SAFA treats it as a relationship between noise and source.

Our contributions are as follows.

\begin{itemize}
    \item We show that noise effectiveness in a frozen conditional generator is dominated by low-rank source-noise interactions, which a lightweight selector suffices to learn.
    \item We propose SAFA, which freezes the generator after conditioning and supervises pre-generation noise selection with the generator's own response matrix, preserving image quality and yielding more usable samples than generator optimization.
    \item We find that prediction before generation matches measurement after generation in cross-encoder similarity, and validate SAFA through cross-encoder evaluation and downstream expression recognition, with a 16.8\% relative accuracy gain.
\end{itemize}

\section{Related Work}
\label{sec:related}

\paragraph{Generator optimization with representation constraints.}
Representation-conditioned diffusion models can generate diverse images that preserve representation information~\cite{ref7}. To move outputs closer to a target representation, a common approach adds a representation similarity loss to the generation loss and coordinates the two gradients by multi-objective optimization, e.g., projecting conflicting gradients (PCGrad~\cite{ref1}), seeking worst-case improvement near the average gradient (CAGrad~\cite{ref2}), or adaptively weighting tasks (FAMO~\cite{ref3}). Another line fine-tunes generators with similarity as a differentiable reward~\cite{ref4,ref8}, often with low-rank adaptation~\cite{ref9}. Over-optimizing a reward model can drive outputs away from the true objective~\cite{ref10}. SAFA trains the generator only during conditioning with its native loss and never updates it through any encoder's gradients.

\paragraph{Noise selection and optimization.}
Initial noise provides another way to control a frozen generator's outputs. Good Seed~\cite{ref5} studies global seed rankings across prompts, while NoiseQuery~\cite{ref6} retrieves noise vectors matching a target through a feature index of unconditionally generated images. Ma et al.~\cite{ref11} consider search in noise space as a dimension of inference-time scaling. ReNO~\cite{ref14} adjusts initial noise through reward optimization for each request; Golden Noise~\cite{ref15} and Noise Hypernetworks~\cite{ref16} turn noise adjustment into an offline-learned mapping. HPM-Predict~\cite{ref12} and PAINE~\cite{ref13} predict preference scores for condition-noise pairs before generation. SAFA targets emotion preservation for a specific source image and learns source-specific selection from the low-rank interactions shared by the complete responses of different sources.

\paragraph{Facial expression data synthesis and augmentation.}
SynFER~\cite{ref17} controls expression generation with text and facial action units and evaluates synthetic data by recognition accuracy. DA-Fusion~\cite{ref18} and DiffuseMix~\cite{ref19} apply diffusion models to semantic augmentation of real images, and LivePortrait~\cite{ref20} transfers expressions across faces. TrivialAugment~\cite{ref21} uses single-image transformations, whereas MixUp~\cite{ref22} and CutMix~\cite{ref23} mix images and labels. SAFA instead preserves the emotion representation of a specific source and obtains multiple usable samples with few generations.

\section{Method}
\label{sec:method}

\begin{figure*}[t]
    \centering
    \includegraphics[width=\textwidth]{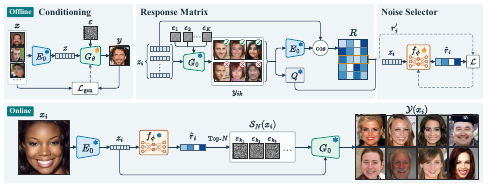}
    \caption{SAFA pipeline. Offline, conditioning with the native generation loss produces a frozen conditional generator $G_0$. Candidate images are then generated for $n_{\text{tr}}$ training sources and $K$ noise vectors. The encoder $E_0$ computes the response matrix $R$, which supervises the noise selector $f_\phi$, with a fixed penalty subtracted from supervision values for candidates that fail quality assessment. Online, $f_\phi$ predicts responses from the source representation and selects the top $N$ noise vectors for generation by $G_0$.}
    \label{fig:pipeline}
\end{figure*}

SAFA comprises three offline stages and one online step (Fig.~\ref{fig:pipeline}). The offline stages are conditioning (Sec.~\ref{sec:pushpick}) and response matrix construction and selector training (Sec.~\ref{sec:selector}), and the online step selects noise before generation. Section~\ref{sec:statistics} analyzes its statistical properties.

\subsection{Problem Definition}
\label{sec:problem}

Given a source image $x$, a frozen representation encoder $E$ produces an emotion representation $z=E(x)\in\mathbb{R}^{d}$, and a conditional generator $G_\theta$ generates a face $G_\theta(z,\varepsilon)$ from $z$ and noise $\varepsilon$. The representation similarity between a generated image $y$ and its source is

\begin{equation}
s_E(y,x)=\cos\big(E(y),\,E(x)\big).
\label{eq:similarity}
\end{equation}

The training similarity $s_0$ is measured by the default encoder $E_0$. We draw $K=128$ noise vectors from $\mathcal{N}(0,I)$ once as the noise dictionary $\mathcal{E}=\{\varepsilon_k\}_{k=1}^{K}$. For each source, the goal is to obtain, within $N=8$ generations, as many images as possible that are similar to the source, pass the quality gate $Q(y)=1$, and differ in appearance. For the $N$ generated images $\mathcal{Y}(x)$, the joint measure of these three requirements is the usable yield

\begin{equation}
\begin{aligned}
&U_N(x)=\max\Big\{\,\lvert T\rvert\;\Big|\;T\subseteq\mathcal{Y}(x),\\
&\quad Q(y)=1,\;s_0(y,x)\ge\tau\quad\forall y\in T,\\
&\quad d(y,y')\ge\delta\quad\forall y\neq y'\in T\Big\},
\end{aligned}
\label{eq:yield}
\end{equation}

where $d$ is the LPIPS distance and $\tau$ and $\delta$ are the similarity and distance thresholds.

\subsection{Generator Optimization and Noise Selection}
\label{sec:pushpick}

The base generator $G_{\text{base}}$ is pretrained on FFHQ~\cite{ref24} and CelebA~\cite{ref25} without emotion annotations. In the conditioning stage, we continue training it on $n_c=98{,}949$ high-quality faces, each conditioned on its own emotion representation, with only the native generation loss $\mathcal{L}_{\text{gen}}$. The encoder only provides the condition and passes no gradient, which yields the conditional generator $G_0=G_{\theta_0}$.

Representation preservation can then be improved in two ways. Generator optimization (push) further updates the parameters to increase representation similarity,

\begin{equation}
\begin{aligned}
\min_{\theta}\quad &\mathcal{L}_{\text{gen}}(\theta)\\
&+\beta\,\mathbb{E}_{x,\varepsilon}\big[1-s_0\big(G_\theta(z,\varepsilon),x\big)\big],
\end{aligned}
\label{eq:push}
\end{equation}

where multi-objective optimization or reward fine-tuning balances the two terms. Noise selection (pick) keeps $\theta_0$ fixed and only decides which noise vectors to use,

\begin{equation}
\mathcal{S}_N(x)=\operatorname*{arg\,top\text{-}N}_{k\in\{1,\dots,K\}}\;\hat{r}_k(z),
\label{eq:selection}
\end{equation}

where $\hat{r}_k(z)$ predicts the quality-adjusted response of source $x$ under noise $\varepsilon_k$. Generator optimization changes the mapping from noise to images, whereas every image chosen by noise selection remains a sample of $G_0$ under standard Gaussian noise.

In our experiments, generator optimization methods start from $G_{\text{base}}$ and jointly train the conditioning adapter and the generator, while noise selection methods share $G_0$ with SAFA, which gives these baselines the benefit of our conditioning stage.

\subsection{Response Matrix and Noise Selector}
\label{sec:selector}

For a training source $x_i$ and noise $\varepsilon_k$, we generate a candidate $y_{ik}=G_0(z_i,\varepsilon_k)$ and record its response

\begin{equation}
r_{ik}=s_0(y_{ik},x_i).
\label{eq:response}
\end{equation}

Each row of the response matrix $R=[r_{ik}]$ holds the actual similarities of one source under all $K$ noise vectors. With $n_{\text{tr}}=1{,}024$ training and $n_{\text{dev}}=128$ development sources, construction takes $(n_{\text{tr}}+n_{\text{dev}})K$ generations.

The noise selector $f_\phi:\mathbb{R}^{d}\to\mathbb{R}^{K}$ maps $z$ to predicted responses $\hat{r}=f_\phi(z)$ for all noise vectors and is trained with

\begin{equation}
\begin{aligned}
\mathcal{L}(\phi)={}&\frac{1}{K}\sum_{k=1}^{K}\big(\hat{r}_{ik}-r_{ik}\big)^2\\
&-\frac{\lambda}{N}\sum_{k\in\mathcal{T}_N(i)}\log\frac{\exp\hat{r}_{ik}}{\sum_{k'=1}^{K}\exp\hat{r}_{ik'}},
\end{aligned}
\label{eq:loss}
\end{equation}

where $\mathcal{T}_N(i)$ indexes the $N$ highest responses in row $i$ and $\lambda=0.5$. The first term fits the whole row, and the second raises the true top $N$ in the predicted ranking, penalizing errors near the top more strongly. During construction we also record the quality assessment of each candidate and subtract a fixed penalty $w=0.2$ from the supervision of failed ones, replacing $r_{ik}$ in Eq.~\eqref{eq:loss} with $r'_{ik}=r_{ik}-w\,(1-Q(y_{ik}))$, so no quality model runs at inference. The selector is a three-layer MLP with about 1.7M parameters, chosen on the development set among MLP, GRU, LSTM, Transformer, and nearest-neighbor retrieval. Inference takes one encoding and one forward pass, after which $G_0$ generates from $\mathcal{S}_N(x)$.

Since the supervision comes from the samples of $G_0$ itself, its ceiling is set by the coverage of the noise dictionary, which a larger dictionary raises (Sec.~\ref{sec:mechanism}).

\subsection{Statistical Analysis of Noise Selection}
\label{sec:statistics}

\paragraph{Interaction effect decomposition.}
Following two-way analysis of variance, we decompose the response matrix as

\begin{equation}
\begin{aligned}
r_{ik}&=\mu+a_i+b_k+c_{ik},\\
\sum_i a_i&=\sum_k b_k=0,\quad
\sum_i c_{ik}=\sum_k c_{ik}=0.
\end{aligned}
\label{eq:decomposition}
\end{equation}

The source main effect $a_i$ reflects how hard a source is to generate regardless of noise, the noise main effect $b_k$ the common effect of a noise vector on all sources, and the interaction effect $c_{ik}$ its source-specific effect. Averaging the within-source variance across noise vectors gives the noise-dependent variance $V_{\text{n}}=\operatorname{Var}(b)+\operatorname{Var}(c)$ and the interaction share $\pi_c=\operatorname{Var}(c)/V_{\text{n}}$. A global noise ranking exploits only $b_k$, whereas source-specific selection can also exploit $c_{ik}$. If the interaction matrix $C=[c_{ik}]$ concentrates its energy in the first $r$ singular directions (high rank-$r$ energy $\eta_r$), the noise preferences of different sources can be expressed as combinations of a few shared directions.

\paragraph{Selection gains.}
For a given source, let $v_k$ be the evaluation similarity under noise $\varepsilon_k$ and $u_k$ the selection signal used for ranking. If $(u,v)$ is approximately bivariate normal, selecting the top $N$ of $M$ candidates by $u$ gives an expected evaluation similarity of~\cite{ref26}

\begin{equation}
\mathbb{E}\big[\bar{v}_{\text{sel}}\big]=\mu_v+\iota(N,M)\,\rho_{uv}\,\sigma_v,
\label{eq:intensity}
\end{equation}

where the selection intensity $\iota(N,M)$ is the expected mean of the largest $N$ of $M$ standard normal samples, $\rho_{uv}$ is the within-source correlation between $u$ and $v$, and $\sigma_v$ is the standard deviation of $v$ across noise vectors, which is set by the generator. The intensity $\iota$ depends only on how many candidates are ranked, not on how many are generated. Best-of-$N$~\cite{ref11} must generate all $M$ candidates, whereas SAFA ranks $K$ candidates and generates only $N$, obtaining $\iota(N,K)$ with $N$ generations ($\iota(8,128)\approx1.94$). The correlation $\rho_{uv}$ measures how well the signal tracks $v$. The signal of Good Seed~\cite{ref5} contains only $b_k$ and that of NoiseQuery~\cite{ref6} comes from unconditional outputs, whereas the signal of SAFA covers the interaction effects, the bulk of the noise-dependent variance.

\paragraph{Measurement error and the optimizer's curse.}
Write the training similarity measured after generation as $u_k=v^{*}_k+e_k$, where $e_k$ is the image-specific error of the encoder. Selecting the largest measured values favors images with both high true similarity and large errors, and the more candidates and the more direct the optimization, the larger the selected errors become, which is the optimizer's curse~\cite{ref27}. Generator optimization is its most direct form, as gradients update parameters toward high encoder scores~\cite{ref10}. The prediction of SAFA is a function of the source representation learned from structure shared across sources, so it averages out image-specific errors, giving lower training similarity than per-image measurement but comparable cross-encoder similarity.

\paragraph{Support invariance.}
Every SAFA output is $y=G_0(z,\varepsilon_k)$ with a standard Gaussian $\varepsilon_k$, so for any source

\begin{equation}
\operatorname{supp}\,p_{\text{SAFA}}(\cdot\mid z)\subseteq\operatorname{supp}\,p_{G_0}(\cdot\mid z).
\label{eq:support}
\end{equation}

Noise selection only reweights samples and never produces images that $G_0$ cannot generate, whereas generator optimization changes $\theta$ and can leave the manifold learned by $G_0$, as examined in Sec.~\ref{sec:mechanism} with manifold precision~\cite{ref28}.

\section{Experiments}
\label{sec:experiments}

\subsection{Experimental Setup}
\label{sec:setup}

\noindent\textbf{Data and models.} Source images come from AffectNet~\cite{ref29}, which has over one million Internet face images, about 450,000 of them manually annotated. We split it into disjoint $n_{\text{tr}}$, $n_{\text{dev}}$, and $n_{\text{te}}=512$ sources and use the official validation set only for downstream testing. All hyperparameters and checkpoints are selected on the development set. The default generator is the one-step MeanFlow B/2~\cite{ref30}, and $E_0$ is ResNet-50~\cite{ref31}.

\noindent\textbf{Evaluation.} Augmented samples are used to train other models, so their emotion information should be readable by encoders not involved in training. Our primary metric is the cross-encoder similarity $\bar{s}=\tfrac12(s_{E_A}+s_{E_B})$, computed with two held-out encoders, $E_A$ (Swin-T~\cite{ref32}) and $E_B$ (MobileNetV3~\cite{ref33}), which are excluded from the training, selection, and tuning of all methods; $s_0$ is reported separately. In Sec.~\ref{sec:generalization}, ConvNeXt-T~\cite{ref34} replaces $E_0$ for conditioning and response matrix construction, with evaluation unchanged. All three encoders are trained only on the eight-class labels of the AffectNet training set, excluding validation images and test sources.

Following ArtifactReward~\cite{ref40}, a frozen Qwen2.5-VL-7B-Instruct assesses each image with identical preprocessing, prompt, and thresholds for all methods, detecting digital artifacts that impair realism, such as lighting anomalies, abnormal object structures or positions, and blending errors. The score is the probability of ``artifact-free,'' normalized over the YES/NO answer groups at the first response token, and $Q(y)=1$ if it exceeds 0.5 with a YES/NO mass of at least 0.5. The pass rate $q$ covers all $n_{\text{te}}N$ outputs, counting abstentions as failures. We do not use FID as a per-image gate because it penalizes the appearance deviation that augmentation seeks. Usable yield uses $\tau=0.80$ and $\delta=0.127$ with LPIPS~\cite{ref35} as $d$.

Two annotators, blinded to method and similarity scores, rated emotion consistency on 400 source-generated pairs stratified by method and similarity, and artifacts on 400 generated images. Among samples meeting the similarity threshold, 95\% were judged to preserve the source emotion, and $Q$ agreed with human artifact judgments on 96\%.

\noindent\textbf{Compared methods.} Each method returns $N$ images per source. Among generator optimization methods, CAGrad~\cite{ref2}, PCGrad~\cite{ref1}, and FAMO~\cite{ref3} coordinate $\mathcal{L}_{\text{gen}}$ and the similarity loss by multi-objective optimization; DRaFT~\cite{ref4} fine-tunes all parameters through the differentiable generation chain with $s_0$ as the reward; AlignProp~\cite{ref8} backpropagates this reward through intermediate denoising outputs; and LoRA~\cite{ref9} uses the objective of DRaFT but updates only low-rank and conditioning adapters. We select the EMA checkpoint with the highest development $s_0$, matching our primary objective, and use $U_N$ for joint evaluation. Noise selection methods share $G_0$ and include random sampling, Best-of-$N$~\cite{ref11} (generating $M$ random images and keeping the top $N$ by $s_0$), Good Seed~\cite{ref5}, and NoiseQuery~\cite{ref6}.

\begin{table*}[t]
\centering
\caption{Comparison of generator optimization and noise selection. Each method outputs $N$ images for each of $n_{\text{te}}$ test sources. Generator optimization methods train from $G_{\text{base}}$, while noise selection methods use $G_0$. The held-out encoders compute $s_{E_A}$ and $s_{E_B}$; $s_0$ is similarity measured by the training encoder. Online time is the mean time per request to return $N$ images, and offline time is the total one-time preparation, including conditioning. Bold indicates the best value among noise selection methods, excluding the evaluator oracle.}
\label{tab:main}
\small
\setlength{\tabcolsep}{5pt}
\renewcommand{\arraystretch}{1.10}
\begin{tabular}{@{}llcccccccc@{}}
\toprule
Method & Starting point & Generations & $s_{E_A}$ $\uparrow$ & $s_{E_B}$ $\uparrow$ & $s_0$ $\uparrow$ & $q$ $\uparrow$ & $U_N$ $\uparrow$ & Online (s) & Offline (h) \\
\midrule
\multicolumn{10}{l}{\emph{Generator optimization}} \\
PCGrad~\cite{ref1} & $G_{\text{base}}$ & $N$ & 0.662 & 0.685 & 0.986 & 0.10\% & 0.01 & 0.15 & 33.4 \\
CAGrad~\cite{ref2} & $G_{\text{base}}$ & $N$ & 0.801 & 0.769 & 0.987 & 6.25\% & 0.25 & 0.15 & 33.0 \\
FAMO~\cite{ref3} & $G_{\text{base}}$ & $N$ & 0.726 & 0.745 & 0.988 & 0.56\% & 0.03 & 0.15 & 37.9 \\
DRaFT~\cite{ref4} & $G_{\text{base}}$ & $N$ & 0.785 & 0.768 & 0.987 & 0.00\% & 0.00 & 0.15 & 24.1 \\
AlignProp~\cite{ref8} & $G_{\text{base}}$ & $N$ & 0.795 & 0.802 & 0.988 & 0.44\% & 0.03 & 0.15 & 19.4 \\
LoRA~\cite{ref9} & $G_{\text{base}}$ & $N$ & 0.664 & 0.641 & 0.749 & 34.72\% & 1.54 & 0.16 & 37.6 \\
\midrule
\multicolumn{10}{l}{\emph{Noise selection}} \\
Random sampling & $G_0$ & $N$ & 0.507 & 0.475 & 0.515 & 67.2\% & 1.87 & 0.15 & 7.7 \\
Best-of-$N$~\cite{ref11} & $G_0$ & $K$ & 0.689 & 0.665 & \textbf{0.908} & 65.9\% & 4.97 & 2.35 & 7.7 \\
Good Seed~\cite{ref5} & $G_0$ & $N$ & 0.632 & 0.599 & 0.702 & 58.7\% & 2.44 & 0.15 & 8.0 \\
NoiseQuery~\cite{ref6} & $G_0$ & $N$ & 0.546 & 0.507 & 0.681 & 61.3\% & 2.42 & 0.16 & 7.9 \\
\textbf{SAFA} & $G_0$ & $N$ & \textbf{0.694} & \textbf{0.670} & 0.885 & \textbf{87.4\%} & \textbf{5.85} & 0.15 & 15.6 \\
Evaluator oracle & $G_0$ & $K$ & 0.853 & 0.850 & 0.738 & 72.4\% & 4.98 & --- & --- \\
\bottomrule
\end{tabular}
\end{table*}

\begin{figure}[t]
\centering
\includegraphics[width=\columnwidth]{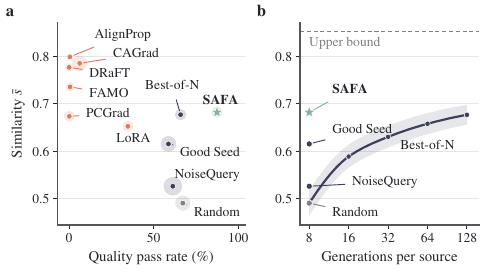}
\caption{Comparison of generator optimization and noise selection. (a) Quality pass rate and cross-encoder similarity. (b) Cross-encoder similarity versus generations per source for noise selection methods. SAFA matches Best-of-$N$ at $M=128$ with 8 generations; the dashed line is the evaluator oracle.}
\label{fig:pushpick}
\end{figure}

\subsection{Comparison with Generator Optimization}
\label{sec:mainresults}

\noindent\textbf{Generator optimization increases representation similarity but sharply reduces usable yield (Table~\ref{tab:main}, Fig.~\ref{fig:pushpick}a).} Except for LoRA, generator optimization methods drive development-set training similarity close to saturation and also raise held-out similarity, mostly exceeding SAFA in cross-encoder similarity. However, their quality pass rates are mostly below 10\%, and their samples show extensive structural artifacts and leave the real-face manifold (Fig.~\ref{fig:manifold}). This indicates that expression encoders of the same kind share correlated measurement errors, so optimization against $E_0$ partially transfers to them but not to quality assessment~\cite{ref10}. Noise selection methods keep quality near the level of $G_0$, and SAFA further raises it together with similarity through the quality penalty (Sec.~\ref{sec:mechanism}). Among them, SAFA attains the highest cross-encoder similarity and 3.1 times the usable yield of random sampling, whereas Good Seed and NoiseQuery, which exploit only the noise main effect or unconditional outputs, give limited gains.

\noindent\textbf{SAFA matches exhaustive selection in representation similarity at a lower generation cost (Fig.~\ref{fig:pushpick}b).} Best-of-$N$ improves steadily with $M$ but matches the $N$-generation result of SAFA only at $M=K$. By Eq.~\eqref{eq:intensity}, both have the same selection intensity, but Best-of-$N$ must first generate all candidates. The evaluator oracle selects the top $N$ of $K$ candidates by $\bar{s}$ itself and serves only as a reference.

\noindent\textbf{SAFA requires less offline cost than generator optimization (Table~\ref{tab:main}).} Its one-time preparation takes 15.6 hours, including 7.7 hours of conditioning shared by all noise selection methods, 2.7 hours of candidate generation, 5.2 hours of quality assessment, and about 18 seconds of selector training, versus 19.4 to 37.9 hours for generator optimization. The response matrix doubles the offline cost of other noise selection methods, but relative to Best-of-$N$ with the same quality penalty this extra cost is recovered within about 1,400 sources (Sec.~\ref{sec:mechanism}).

\subsection{Facial Expression Recognition with Scarce Data}
\label{sec:downstream}

To simulate scarce and imbalanced rare-emotion data, we sample $n_s=1{,}200$ labeled images by class (50 to 300 per class) from the AffectNet training set to form the scarce subset AffectNet-S, train ResNet-18~\cite{ref31} and ConvNeXt-T~\cite{ref34} classifiers on it, and test on the official validation set. All augmentation methods use the same budget and training steps, and generative methods add the same number of images, generated only from training sources.

\noindent\textbf{SAFA's generated samples provide the largest improvement for both classifiers (Table~\ref{tab:recognition}),} raising accuracy over real images alone by 16.8\% and 17.7\% in relative terms. SAFA also outperforms random generation filtered by the same quality and distinctness criteria, so this extra gain comes from noise selection itself. Conventional augmentation helps little with scarce data, and LivePortrait~\cite{ref20}, which transfers source expressions to other real faces, brings no gain on ResNet-18, suggesting that preserving the source emotion representation matters more than introducing new real appearances.

\begin{table}[t]
\centering
\caption{Facial expression recognition accuracy (\%) on AffectNet-S. Mean $\pm$ standard deviation over five random seeds.}
\label{tab:recognition}
\small
\setlength{\tabcolsep}{3pt}
\renewcommand{\arraystretch}{1.10}
\begin{tabular}{@{}lcc@{}}
\toprule
Augmentation method & ResNet-18 & ConvNeXt-T \\
\midrule
Real images only & 34.0 $\pm$ 0.6 & 36.8 $\pm$ 1.3 \\
+ TrivialAugment~\cite{ref21} & 35.3 $\pm$ 1.1 & 37.5 $\pm$ 2.2 \\
+ MixUp~\cite{ref22} & 35.0 $\pm$ 0.4 & 38.9 $\pm$ 0.6 \\
+ CutMix~\cite{ref23} & 33.8 $\pm$ 1.0 & 35.5 $\pm$ 1.5 \\
+ LivePortrait~\cite{ref20} & 33.2 $\pm$ 1.2 & 38.8 $\pm$ 1.1 \\
+ Random generation & 36.4 $\pm$ 0.8 & 40.7 $\pm$ 0.8 \\
\textbf{+ SAFA} & \textbf{39.7 $\pm$ 0.4} & \textbf{43.3 $\pm$ 1.2} \\
\bottomrule
\end{tabular}
\end{table}

\subsection{Generalization under Component Replacement}
\label{sec:generalization}

We replace the encoder, generator, or dataset one at a time (Fig.~\ref{fig:pipeline}), reconstructing the response matrix and retraining the selector each time. The search gain ratio $\gamma=(\bar{s}_{\text{SAFA}}-\bar{s}_{\text{rand}})/(\bar{s}_{\text{BoN-}K}-\bar{s}_{\text{rand}})$ is the fraction of the Best-of-$N$ gain that SAFA attains with $N$ generations, normalizing absolute similarity across components for direct comparison.

\noindent\textbf{SAFA's gains remain consistent across component replacements (Table~\ref{tab:components}).} SAFA improves cross-encoder similarity in every setting, including ConvNeXt-T as the encoder and retraining on RAF-DB~\cite{ref36}, so it depends on neither a particular encoder architecture nor a particular dataset. Absolute gains vary across generators~\cite{ref37,ref38,ref39}. SDXL has the smallest within-source variation $\sigma_v$ (0.223, 0.252, 0.213, and 0.126 for MeanFlow B/2, LDM, LCM, and SDXL, respectively) and the smallest gain, consistent with the decrease in gain with $\sigma_v$ in Eq.~\eqref{eq:intensity}.

\begin{table}[t]
\centering
\caption{SAFA's cross-encoder similarity $\bar{s}$ when replacing one component at a time, with $N$ images per source. The default setting uses AffectNet, MeanFlow B/2, and ResNet-50. Bold identifies SAFA rather than the best value.}
\label{tab:components}
\footnotesize
\setlength{\tabcolsep}{2.3pt}
\renewcommand{\arraystretch}{1.13}
\begin{tabular}{@{}>{\raggedright\arraybackslash}p{0.16\columnwidth}>{\raggedright\arraybackslash}p{0.25\columnwidth}cccc@{}}
\toprule
\shortstack[l]{Replaced\\component} & Setting & Random & Best-of-$N$ & SAFA & $\gamma$ \\
\midrule
None & Default & 0.491 & 0.677 & \textbf{0.682} & 102.7\% \\
Dataset & RAF-DB~\cite{ref36} & 0.579 & 0.714 & \textbf{0.711} & 97.7\% \\
Encoder & ConvNeXt-T~\cite{ref34} & 0.519 & 0.743 & \textbf{0.736} & 96.9\% \\
Generator & LDM~\cite{ref37} & 0.387 & 0.645 & \textbf{0.645} & 99.8\% \\
Generator & LCM~\cite{ref38} & 0.393 & 0.593 & \textbf{0.592} & 99.7\% \\
Generator & SDXL~\cite{ref39} & 0.735 & 0.812 & \textbf{0.776} & 53.4\% \\
\bottomrule
\end{tabular}
\end{table}

\subsection{Mechanism Analysis and Ablations}
\label{sec:mechanism}

\noindent\textbf{Roles of conditioning and noise selection (Table~\ref{tab:conditioning}).} Conditioning establishes an interface for emotion representations, moving cross-encoder similarity from significantly below to significantly above the unrelated-pair reference ($p<10^{-40}$ and $p<10^{-50}$, respectively) while keeping the pass rate above 67\%; all noise selection methods share this stage. On top of it, SAFA improves cross-encoder similarity, training similarity, and pass rate by 39\%, 72\%, and 30\% in relative terms.

\noindent\textbf{The quality penalty raises the pass rate with little effect on similarity (Table~\ref{tab:conditioning}).} Without it ($w=0$), the pass rate of SAFA drops from 87.4\% to 66.8\%, on par with random sampling, while cross-encoder similarity remains essentially unchanged. With the same penalty, Best-of-$N$ attains a higher pass rate and usable yield, since per-image assessment directly excludes images with artifacts, but lower cross-encoder similarity, and each request must generate and assess all $K$ candidates online in 20.4 s, about 136 times the time of SAFA.

\begin{table}[t]
\centering
\caption{Roles of conditioning, noise selection, and the quality penalty $w$. Best-of-$N$ ranks the $K$ generated candidates by $r'$. $n_{\text{te}}$ test sources, with $N$ images per source. Bold indicates the best value in each column.}
\label{tab:conditioning}
\footnotesize
\setlength{\tabcolsep}{3pt}
\renewcommand{\arraystretch}{1.10}
\begin{tabular}{@{}lccccc@{}}
\toprule
Configuration & $s_{E_A}$ $\uparrow$ & $s_{E_B}$ $\uparrow$ & $s_0$ $\uparrow$ & $q$ $\uparrow$ & $U_N$ $\uparrow$ \\
\midrule
$G_{\text{base}}$ + Random sampling & 0.160 & 0.138 & 0.023 & 68.9\% & 0.00 \\
$G_0$ + Random sampling & 0.507 & 0.475 & 0.515 & 67.2\% & 1.87 \\
$G_0$ + Best-of-$N$ ($w=0$) & 0.689 & 0.665 & \textbf{0.908} & 65.9\% & 4.97 \\
$G_0$ + Best-of-$N$ ($w=0.2$) & 0.685 & 0.659 & 0.892 & \textbf{95.9\%} & \textbf{6.79} \\
$G_0$ + SAFA ($w=0$) & 0.688 & 0.665 & 0.882 & 66.8\% & 4.71 \\
$G_0$ + SAFA ($w=0.2$) & \textbf{0.694} & \textbf{0.670} & 0.885 & 87.4\% & 5.85 \\
\bottomrule
\end{tabular}
\end{table}

\begin{figure}[t]
\centering
\includegraphics[width=\columnwidth]{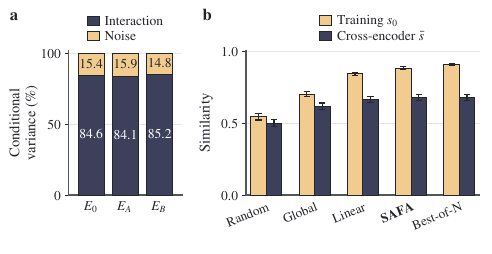}
\caption{Response matrix structure and decomposition of selection signals. (a) Two-way variance decomposition of the response matrices of $E_0$, $E_A$, and $E_B$. (b) Cross-encoder similarity (dark blue) and training similarity (light yellow) as the selection signal is progressively strengthened on the same set of $K$ candidates; SAFA generates only $N$ images.}
\label{fig:mechanism}
\end{figure}

\begin{figure}[t]
\centering
\includegraphics[width=\columnwidth]{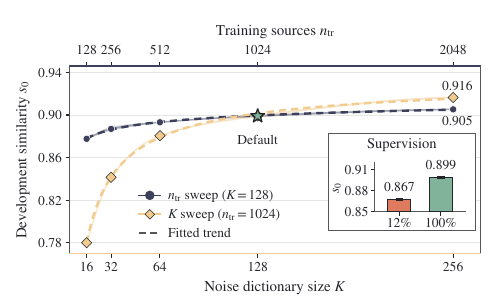}
\caption{Effect of response matrix scale on selection performance. The dark blue curve uses the upper horizontal axis and increases the number of training sources with $K=128$ fixed. The light yellow curve uses the lower horizontal axis and increases the noise dictionary size with $n_{\text{tr}}=1{,}024$ fixed. Stars mark the default setting, and dashed lines show fitted trends. The lower-right inset compares supervision on only 12\% of candidate entries with complete response supervision.}
\label{fig:scaling}
\end{figure}

\begin{figure}[t]
\centering
\includegraphics[width=\columnwidth]{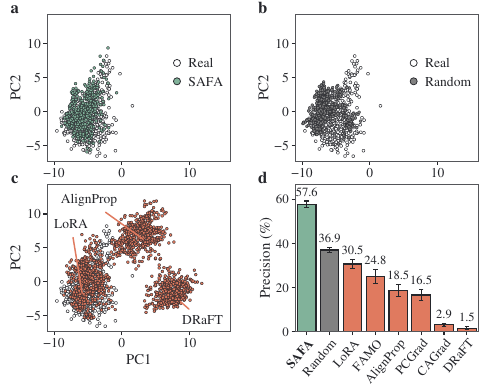}
\caption{Noise selection preserves the generator's output manifold. (a) SAFA, (b) random sampling, and (c) AlignProp, LoRA, and DRaFT. (d) Manifold precision $P_{\text{man}}$ for each method.}
\label{fig:manifold}
\end{figure}

\begin{figure*}[t]
\centering
\includegraphics[width=\textwidth]{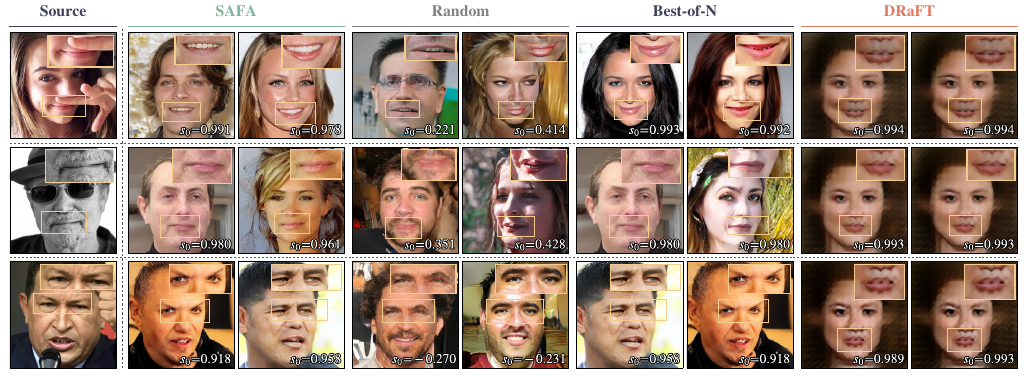}
\caption{Emotion-preserving augmentation across sources. Each row corresponds to one source. From left to right: the source image, SAFA, random sampling, Best-of-$N$, and DRaFT. Representation similarity $s_0$ is shown at the lower right of each image.}
\label{fig:cases}
\end{figure*}

\noindent\textbf{The effectiveness of noise depends primarily on its match with the source (Fig.~\ref{fig:mechanism}a).} The interaction share $\pi_c$ is about 85\% for $E_0$, $E_A$, and $E_B$, and the noise main effect only about 15\%, so any global noise ranking exploits at most a small fraction of the noise-dependent variance, making source-specific selection necessary. The interaction matrix is also low-rank, with $\eta_r\approx70\%$ ($r=8$), so similar sources prefer similar noise and interactions are predictable from source representations. The dominant interaction subspaces of the three encoders overlap substantially ($\omega=\|A^\top B\|_F^2/r$, where $A$ and $B$ are orthonormal bases of two interaction subspaces), about 7 to 12 times the random baseline $r/K$, suggesting that this structure arises mainly from the generator-source pairing rather than any single encoder.

\noindent\textbf{Learnable selection signals concentrate in low-rank interaction effects (Fig.~\ref{fig:mechanism}b).} We progressively strengthen the selection signal on the same $K$ candidates to separate its components. On the development set, the noise main effect alone recovers 65.3\% of the per-image measurement gain, and adding the linear component of the leading $r$ interaction directions raises it to 91.4\%, with the bootstrap confidence interval for every increment excluding zero. A lightweight selector can thus learn these signals. SAFA recovers 99.7\% and surpasses linear low-rank filtering, suggesting a benefit from its nonlinear mapping, and matches per-image measurement in cross-encoder similarity despite lower training similarity, consistent with fitting less of the error specific to $E_0$.

\noindent\textbf{Response matrix scale and cost (Fig.~\ref{fig:scaling}).} Increasing $n_{\text{tr}}$ improves the prediction accuracy $\rho_{uv}$, and increasing $K$ raises the selection intensity $\iota(N,K)$. Both yield continued but diminishing gains per doubling, whereas the offline construction cost grows linearly with $n_{\text{tr}}K$ and the online cost stays at $N$ generations. We therefore use $n_{\text{tr}}=1{,}024$ and $K=128$. Complete supervision is essential, as supervising only 12\% of the entries lowers development $s_0$ from 0.899 to 0.867.

\noindent\textbf{Noise selection preserves the generator's output manifold (Fig.~\ref{fig:manifold}).} Following~\cite{ref28}, we estimate the real-face manifold by $k$-nearest neighbors of real-face features and compute manifold precision $P_{\text{man}}$ on 4,096 generated samples per method. SAFA raises $P_{\text{man}}$ from 36.9\% (random sampling from $G_0$) to 57.6\%, consistent with Eq.~\eqref{eq:support}, since it only reweights samples of the generator and favors those closer to real faces. Generator optimization methods reduce $P_{\text{man}}$ to 1.5\% to 30.5\%, in line with their low quality pass rates.

\noindent\textbf{Visual comparison across sources (Fig.~\ref{fig:cases}).} The smile in the first row appears readily even under random sampling, whereas the brow and eye details in the last two rows reveal the role of source-specific selection. SAFA preserves these features with varied appearances, Best-of-$N$ reaches similar results by generating and then filtering, and the high similarity of DRaFT comes with blurred and increasingly similar faces.

\subsection{Limitations}
\label{sec:limitations}

SAFA can only select among samples that the generator produces within the noise dictionary, so its effectiveness grows with generator capability. When the target data differ substantially in imaging style from the conditioning faces, the response matrix should be rebuilt on the target data instead of reusing a selector trained elsewhere. Conditioning and response matrix construction are one-time offline costs, per generator and per target dataset, respectively. Generated samples partly retain source identity, so identity-anonymizing augmentation and explicit diversity control remain future work.

\section{Conclusion}
\label{sec:conclusion}

We study emotion-preserving face augmentation and find that a frozen conditional generator already produces samples highly similar to the source emotion representation, with noise effectiveness determined mainly by low-rank source-noise interactions. Exploiting this structure, SAFA stops optimizing the generator after conditioning and trains a lightweight selector on the generator's own response matrix to choose noise before generation. It raises the quality pass rate, achieves consistent cross-encoder similarity with few generations, and improves expression recognition under scarce labels. A natural next step is a general noise selector trained on response matrices from multiple encoders, generators, and datasets.

\FloatBarrier
\label{page:main-end}
\clearpage
\section*{LLM Usage Statement}
ChatGPT, Codex, and Claude were used for research discussions, code assistance, and manuscript feedback. Qwen2.5-VL-7B-Instruct is used for the image-quality evaluation described in Section~\ref{sec:setup}. The authors retain responsibility for the study, results, and final manuscript.

{
  \small
  \bibliographystyle{ieeenat_fullname}
  \bibliography{refs,supplement_refs}
}

\clearpage
\setcounter{section}{0}
\setcounter{subsection}{0}
\setcounter{figure}{0}
\setcounter{table}{0}
\setcounter{equation}{0}
\renewcommand{\thesection}{\Alph{section}}
\renewcommand{\thefigure}{S\arabic{figure}}
\renewcommand{\thetable}{S\arabic{table}}
\renewcommand{\theequation}{S\arabic{equation}}
\renewcommand{\theHsection}{supp.\Alph{section}}
\renewcommand{\theHfigure}{supp.\arabic{figure}}
\renewcommand{\theHtable}{supp.\arabic{table}}
\renewcommand{\theHequation}{supp.\arabic{equation}}
\graphicspath{{supplement_figures/}}
\newcommand{\safa}{SAFA}
\twocolumn[{\centering\Large\bfseries Supplementary Material\par\vspace{12pt}}]

\noindent This supplement provides implementation and evaluation details, response structure and selector analyses, computational costs, recognition details under scarce expression labels, and additional qualitative comparisons. Figures and tables use an S prefix.

\section{Implementation Details}
\label{sup:sec:implementation}

\paragraph{Conditional generator.}

The default backbone is the one-step MeanFlow B/2 generator~\cite{ref30}. We continue training on 98,949 high-quality training images from FFHQ and CelebA, while holding out 1,024 separate canonical image groups for generator evaluation (distinct from the 1,024 selector training sources). Original images, horizontal flips, and known duplicates are assigned to the same group before splitting the data. Each training draw samples an image uniformly and selects its original or horizontally flipped view with equal probability; the training latent is sampled from that view's cached VAE posterior with a latent scaling factor of $0.18215$.

The frozen encoder $E_0$ provides an $\ell_2$-normalized, 512-dimensional representation of the same target view as the condition. A $512\!\rightarrow\!768\!\rightarrow\!768$ MLP with a SiLU hidden activation maps this representation to an increment added to the native time, horizon, and null-label embedding; the final layer has zero-initialized weights and biases. Conditioning dropout zeros the entire increment with probability $0.1$. The generator $G$ and this conditioning adapter are optimized jointly using only the native MeanFlow objective; $E_0$ and the VAE remain frozen.

\begin{table}[t]
\centering

\caption{Default implementation settings. Conditioning trains $G$ and its conditioning adapter; selector training updates only $f_\phi$.}
\label{sup:tab:hyperparameters}
\small
\setlength{\tabcolsep}{3pt}
\renewcommand{\arraystretch}{1.10}
\begin{tabularx}{\linewidth}{@{}lX@{}}
\toprule
Component & Setting \\
\midrule
Image / latent & $256\!\times\!256$ RGB / $4\!\times\!32\!\times\!32$ \\
Conditioning & Adam; batch 256; 20,000 updates; FP32 \\
Conditioning Adam parameters & $(\beta_1,\beta_2)=(0.9,0.95)$; $\epsilon=10^{-8}$; no weight decay \\
Conditioning learning rate & 500-update warmup to $3\!\times\!10^{-5}$; cosine decay to $3\!\times\!10^{-6}$ \\
EMA decay & $0.999$ \\
Response sources & 1,024 training; 128 development \\
Noise dictionary & $K=128$ fixed standard Gaussian tensors \\
Selector & LayerNorm(512); Linear(512, 1024); SiLU; Linear(1024, 1024); SiLU; Linear(1024, 128) \\
Selector optimizer & AdamW; learning rate $0.002$; weight decay $10^{-5}$; $(0.9,0.999)$; $\epsilon=10^{-8}$ \\
Selector schedule & Full batch (1,024); 3,000-update training budget; cosine decay; dev evaluation every 50 updates \\
Online output & Stable top-$8$ indices; one-step generation \\
\bottomrule
\end{tabularx}
\end{table}

\paragraph{Response supervision and selector training.}

We use disjoint AffectNet source sets for selector training ($n_{\mathrm{tr}}=1{,}024$), development ($n_{\mathrm{dev}}=128$), and testing ($n_{\mathrm{te}}=512$). Each training or development source is paired with all $K=128$ fixed noise vectors in $\mathcal{E}$ to construct the complete response matrix. Decoded images are clamped to $[-1,1]$, quantized to 8-bit RGB PNGs, and evaluated on these quantized pixels to obtain $r_{ik}=s_0(y_{ik},x_i)$ and the quality gate $Q(y_{ik})$. Using $w=0.2$ and $\lambda=0.5$, the selector $f_\phi$ (1,707,136 parameters) is trained to predict $r'_{ik}=r_{ik}-w(1-Q(y_{ik}))$ via the main-paper objective in Eq.~(6). Both the regression targets and ranking supervision in Eq.~(6) use the quality-adjusted responses $r'_{ik}$; specifically, $\mathcal{T}_N(i)$ contains the $N=8$ noise indices with the highest $r'_{ik}$ for source $i$. We select the checkpoint with the highest mean measured quality-adjusted response $r'$ among its predicted top-eight candidates on the development set, choosing the earlier checkpoint in case of ties.

\paragraph{Generator optimization baselines.}

All six generator optimization baselines initialize from the native pretrained MeanFlow B/2 backbone with a fresh conditioning adapter, differentiating $1-s_0$ through the generated image, frozen VAE, and frozen $E_0$. CAGrad~\cite{ref2} uses $\alpha=0.4$ with a simplex-constrained SLSQP solver; PCGrad~\cite{ref1} projects conflicting task gradients onto the hyperplane orthogonal to the other task's original gradient and sums the results after logical-batch accumulation; FAMO~\cite{ref3} uses a weight learning rate of $0.025$, a regularization coefficient of $0.01$, and pre-/post-update losses on the same batch; DRaFT~\cite{ref4} updates the full backbone and conditioning MLP through the differentiable one-step generation chain; and LoRA~\cite{ref9} freezes the pretrained backbone and trains rank-$8$, scale-$8$ adapters in the attention $qkv$ and output projections, together with the conditioning MLP. These five methods use the native generation loss and $1-s_0$ as their two objectives: CAGrad, PCGrad, and FAMO combine them using their respective update rules, whereas DRaFT and LoRA optimize their unweighted sum. All five use Adam with batch size 32, gradient clipping at $1$, EMA decay $0.999$, a 500-update warmup to $3\!\times\!10^{-5}$, and cosine decay to $3\!\times\!10^{-6}$.

AlignProp on B/2~\cite{ref8} trains rank-$4$ adapters in the separate query, key, value, and output projections, together with the conditioning MLP, using the reward-only loss $0.01(1-s_0)$. It uses AdamW with learning rate $10^{-3}$, weight decay $0.1$, gradient clipping at $5$, batch size 32, and EMA decay $0.999$.

For the six joint runs from $G_{\mathrm{base}}$, we evaluate the EMA weights every 500 updates on 128 development sources using eight fixed noise vectors, and select the checkpoint with the highest mean $s_0$, choosing the earlier one in case of ties (see Sec.~D.3 for the warm-started runs from $G_0$).

\paragraph{Online timing protocol.}

Online latencies in main-paper Table~1 are averaged over 48 measured requests after two warm-up requests (returning eight images each) on a single NVIDIA A100 GPU, excluding cold model loading, CPU image decoding, and disk writes, while including source encoding via $E_0$.

A \safa{} request comprises one source encoding, one selector forward pass (0.55 ms), and eight one-step generations, taking 0.1535 s (0.15 s in main-paper Table~1). Unpenalized Best-of-$N$ generates and encodes all $K=128$ candidates before ranking, taking 2.3480 s per request (2.35 s in main-paper Table~1). With the quality penalty, Best-of-$N$ additionally assesses the candidates using the frozen quality model, increasing the reported request latency to 20.4 s. \safa{} invokes no quality model online.

\section{Evaluation Protocols}
\label{sup:sec:evaluation}

\paragraph{Encoders and image inputs.}

The conditioning and response encoder $E_0$ is ResNet-50, while the held-out evaluation encoders $E_A$ and $E_B$ are Swin-T and MobileNetV3, respectively. All three are trained on the eight AffectNet expression classes, excluding the official validation images and test sources. The held-out encoders are kept frozen and used only for evaluation. For any encoder $E$, we compute the cosine similarity between its $\ell_2$-normalized embeddings of the source and generated images, evaluating generated images from saved 8-bit RGB PNGs. For $E_0$, the embedding is the $\ell_2$-normalized output of the 512-dimensional linear projector preceding the eight-class classifier, with RGB inputs resized to $224\!\times\!224$ via antialiased bilinear interpolation and standardized with ImageNet channel statistics; $E_A$ and $E_B$ use their fixed expression preprocessing at $224\!\times\!224$ resolution.

For source $i$, let $y_{i1},\ldots,y_{i8}$ be the returned outputs. The source-level mean-eight score is
\begin{equation}
 \begin{aligned}
 m_E(i)&=\frac{1}{8}\sum_{j=1}^{8}s_E(y_{ij},x_i),\\
 m_{\mathrm{cross}}(i)&=\frac{m_{E_A}(i)+m_{E_B}(i)}{2}.
 \end{aligned}
 \label{sup:eq:mean-eight}
\end{equation}

For the default AffectNet experiment, we average equally over all 512 test sources and report training-encoder similarity $s_0$ and cross-encoder similarity $\bar{s}=(s_{E_A}+s_{E_B})/2$ separately. All eight outputs per source enter these means regardless of their quality assessments.

\paragraph{Frozen artifact assessment.}

We assess artifacts using frozen Qwen2.5-VL-7B-Instruct with the detection prompt released by ArtifactReward~\cite{ref40}:
\begin{quote}\small
Analyze the images to identify any unintentional digital artifacts, concentrating on irregular lighting, object placement, blending errors, or anomalies that could affect realism. Disregard any deliberate artistic styles or intentional surreal elements. Respond with YES if artifacts are detected; if not, respond with NO.
\end{quote}

Each image is converted to RGB, JPEG-encoded at quality 95, decoded, and passed to the model processor. The user message places the prompt before the image, and the chat template includes the assistant generation prefix. A single BF16 forward pass yields the logits at the first answer position; the probabilities below are computed in FP32.

Let $\ell_v$ be the logit for vocabulary token $v$. The YES group comprises \texttt{`` Yes''}, \texttt{``Yes''}, \texttt{``yes''}, \texttt{`` yes''}, \texttt{``YES''}, and \texttt{`` YES''}; the NO group comprises \texttt{`` no''}, \texttt{``no''}, \texttt{`` No''}, \texttt{``No''}, \texttt{`` NO''}, \texttt{``NO''}, and tab-prefixed \texttt{``no''}. Each group's probability mass is computed via log-sum-exp within the group and normalized over the full vocabulary $V$:
\begin{equation}
 \begin{aligned}
 p_C&=\frac{\sum_{v\in C}\exp\ell_v}{\sum_{v\in V}\exp\ell_v},\\
 a&=p_{\mathrm{NO}}+p_{\mathrm{YES}},\qquad
 q_{\mathrm{free}}=\frac{p_{\mathrm{NO}}}{a}.
 \end{aligned}
 \label{sup:eq:quality-probabilities}
\end{equation}

The binary quality gate is
\begin{equation}
 Q(y)=\mathbf{1}\{a\geq0.5\ \text{and}\ q_{\mathrm{free}}>0.5\}.
 \label{sup:eq:quality-gate}
\end{equation}

For the default AffectNet experiment, the quality pass rate $q$ is the mean of $Q$ over all $512\!\times\!8$ outputs per method (in percent). Cases with insufficient YES/NO probability mass or tied conditional probabilities count as failures and remain in the denominator.

\paragraph{Appearance diversity and usable yield.}

Appearance distances use frozen LPIPS~\cite{ref35} with a SqueezeNet backbone on RGB images in $[-1,1]$, squaring channel-normalized feature differences, weighting by learned LPIPS weights, and averaging spatially (FP32 features, FP64 reduction) with threshold $\delta=0.127$. Let $\mathcal{P}_i$ be the family of output subsets whose distinct pairs all satisfy $d(y,y')\geq\delta$, and let $\mathcal{A}_i$ be the outputs satisfying $Q(y)=1$ and $s_0(y,x_i)\geq\tau=0.80$. Then
\begin{align}
 D_8(i)&=\max_{T\in\mathcal{P}_i}|T|,
 \label{sup:eq:distinct-count}\\
 U_8(i)&=\max_{T\in\mathcal{P}_i,\ T\subseteq\mathcal{A}_i}|T|.
 \label{sup:eq:usable-count}
\end{align}

Both maximum-cardinality problems are solved via exhaustive subset enumeration. $D_8$ counts pairwise-distinct outputs irrespective of quality or similarity, whereas $U_8$ counts outputs satisfying all three requirements jointly. We report both metrics averaged over test sources, in images per source, with $D_8\in[1,8]$ and $U_8\in[0,8]$. For \safa{}, unpenalized Best-of-$N$, and random sampling, $D_8=8$ throughout, so differences in $U_8$ among them at this threshold reflect joint passage of the quality and similarity gates.

\paragraph{Similarity alignment in usable yield.}

The similarity gate of $U_8$ applies $s_0\geq\tau$, the same training-encoder score ranked by unpenalized Best-of-$N$, making $U_8$ structurally aligned with $E_0$. Among noise-selection methods, unpenalized Best-of-$N$ attains the highest $s_0$ ($0.908$) and penalized Best-of-$N$ the highest $U_8$ ($6.79$). The optimizer's curse analysis in Sec.~3.4 of the main paper motivates evaluation beyond the training encoder. On the held-out encoders $E_A$ and $E_B$, which provide an independent assessment of emotion-representation similarity, \safa{} achieves $0.694$ and $0.670$ with only $N=8$ generations, exceeding both unpenalized Best-of-$N$ ($0.689$ and $0.665$) and penalized Best-of-$N$ ($0.685$ and $0.659$) that generate all $128$ candidates.

\paragraph{Selection gains and evaluation references.}

Selection-gain recovery is the ratio of aggregate cross-encoder gains,
\begin{equation}
 \gamma=\frac{\bar{s}_{\text{SAFA}}-\bar{s}_{\text{rand}}}
 {\bar{s}_{\text{BoN-}K}-\bar{s}_{\text{rand}}},
 \label{sup:eq:gain-recovery}
\end{equation}

using mean-eight cross-encoder scores, with unpenalized Best-of-$N$ over the same $K$-candidate pool in the denominator and a shared random-sampling reference. Unpenalized Best-of-$N$ generates the full dictionary and selects eight outputs by $s_0$; with the quality penalty it ranks by $r'=r-w\,(1-Q)$ and must evaluate the quality model online, incurring 20.4 s per request. The unconditioned baseline $G_{\text{base}}$ with random sampling yields mean $s_0=0.023$ and usable yield $U_8=0$. The evaluator oracle ranks candidates by $\bar{s}$ and selects a common set of eight images for both held-out encoders, providing an offline upper-bound reference for mean-eight cross-encoder similarity within the same candidate pool.

\paragraph{Duplication and identity check.}

We assess output duplication using within-source pixel comparisons and LPIPS distances, and source--output identity similarity using frozen ArcFace~\cite{supbib_arcface} across the 512 test sources. Among all 14,336 within-source pairs per method, neither exact pixel duplicates nor pairs below the LPIPS threshold $\delta=0.127$ appear in \safa{}, unpenalized Best-of-$N$, or random sampling. For identity evaluation, we use the \texttt{w600k\_r50.onnx} recognition weights and \texttt{det\_10g.onnx} face detector from InsightFace's \texttt{buffalo\_l} model pack, align detected faces using five landmarks to $112\times112$ crops, and compute the cosine similarity between $\ell_2$-normalized embeddings. Averaging valid source--output ArcFace cosine similarities within each source and then equally across the 512 sources yields near-zero means: $0.0089$ for \safa{} (4,096/4,096 valid pairs), $0.0074$ for Best-of-$N$ (4,096/4,096), and $0.0064$ for random sampling (4,094/4,096)---indicating low average identity similarity between the generated faces and their source images.

\section{Response Structure and Selector Analysis}
\label{sec:supp_analysis}

\subsection{Common feature-space projection}

\begin{figure*}[t]
  \centering
  \includegraphics[width=\textwidth]{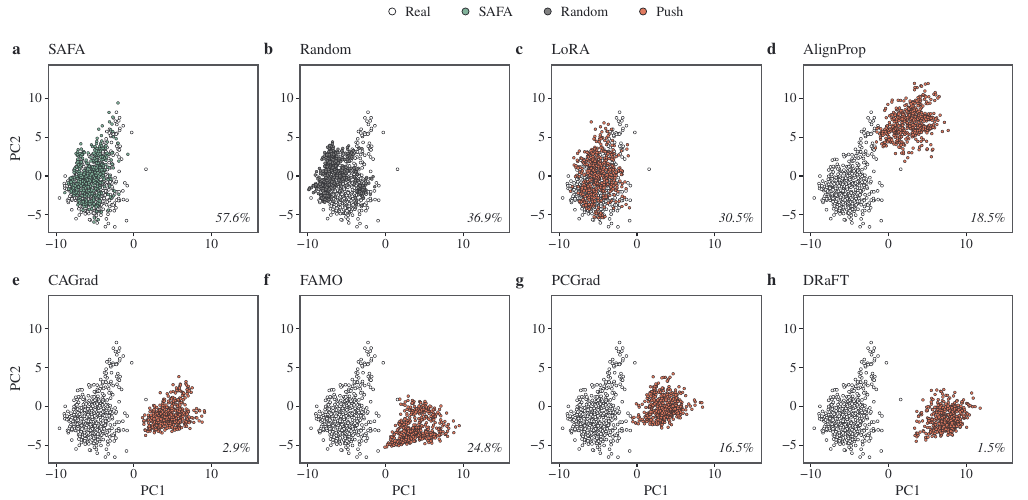}

  \caption{\textbf{Common feature-space projection across methods.} Each panel displays 512 generated samples and the same 512 real-reference samples in a shared PCA space of Inception-2048 features. Bottom-right percentages report manifold precision $P_{\mathrm{man}}$, computed in the original feature space using all 4,096 generated samples per method (main-paper Fig.~6d).}
  \label{fig:supp_projection}
\end{figure*}

Figure~\ref{fig:supp_projection} extends the manifold views in the main paper to all six generator optimization methods (CAGrad, FAMO, PCGrad, DRaFT, LoRA, and AlignProp), \safa{}, and random sampling. Each panel projects 512 generated samples alongside the same 512 real-reference images (hollow gray circles) onto a shared PCA basis of Inception-2048 features with identical axis limits, allowing direct comparison of location and spread. The shared PCA basis is fitted jointly to these 512 real-reference samples and 512 generated samples from each of the eight methods; for each method, the 512 displayed samples comprise all eight outputs from the same 64 test sources selected using a fixed random seed. Samples from \safa{} and random sampling overlap the real-reference distribution, whereas generator optimization methods exhibit varying degrees of distribution shift or concentration in the shared projection. This visualization complements the manifold precision~\cite{ref28} computed in the original feature space and reported in the main paper: manifold precision uses Euclidean distances in the original 2048-dimensional Inception feature space with $k=3$, where each real-reference point defines a ball whose radius is its distance to its third-nearest other real-reference point.

\subsection{Total variance and interaction energy}

\begin{figure*}[t]
  \centering
  \includegraphics[width=\textwidth]{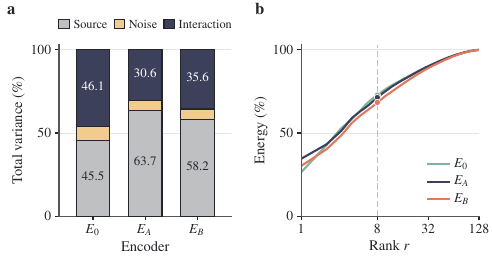}

  \caption{\textbf{Full response decomposition and low-rank interaction structure.} (a) Decomposition of total response variance. (b) Cumulative interaction energy versus rank.}
  \label{fig:supp_structure}
\end{figure*}

For a response matrix $R=[r_{ik}]$, write $r_{ik}=\mu+a_i+b_k+c_{ik}$, where $a_i$ is the source main effect, $b_k$ is the noise main effect, and $c_{ik}$ is the source--noise interaction. Figure~\ref{fig:supp_structure}(a) reports all three components as percentages of the total response variance under $E_0$, $E_A$, and $E_B$. The main paper normalizes by the noise-dependent variance $\operatorname{Var}(b)+\operatorname{Var}(c)$, whereas this figure uses the total response variance, including the source main effect reflecting source difficulty; for selection within a fixed source, only the noise main effect and its interaction with the source govern candidate ranking. Under the main-paper normalization, interactions account for $84.1\%$--$85.2\%$ across the three encoders, highlighting the dominant role of source-specific noise preferences.

Figure~\ref{fig:supp_structure}(b) shows the cumulative singular-value energy of the interaction matrix $C=[c_{ik}]$ on a logarithmic rank axis, with a dashed line and filled markers at $r=8$:
\begin{equation}
  \eta_r=\frac{\sum_{j=1}^{r}\sigma_j(C)^2}{\sum_j\sigma_j(C)^2}.
\end{equation}

The leading directions capture about 70\% of the interaction energy at $r=8$. This concentration motivates a compact predictor of source-specific noise preferences from source representations.

\paragraph{Interaction estimates and subspace overlap.}

The decomposition of Figure~\ref{fig:supp_structure} uses the estimators
\begin{equation}
 a_i=\bar r_{i\cdot}-\mu,\qquad
 b_k=\bar r_{\cdot k}-\mu,\qquad
 c_{ik}=r_{ik}-\bar r_{i\cdot}-\bar r_{\cdot k}+\mu,
 \label{sup:eq:estimators}
\end{equation}

where $\mu$ is the grand mean, and $\bar r_{i\cdot}$ and $\bar r_{\cdot k}$ are the row and column means of $R$. Within sources, the noise-dependent variance contributions and interaction share are
\begin{equation}
 \begin{aligned}
 V_b&=\frac{1}{K}\sum_k b_k^2,\qquad
 V_c=\frac{1}{nK}\sum_i\sum_k c_{ik}^2,\\
 \pi_c&=\frac{V_c}{V_b+V_c}\approx85\%.
 \end{aligned}
 \label{sup:eq:variance-share}
\end{equation}

For orthonormal bases $A,B\in\mathbb{R}^{K\times r}$ of two interaction subspaces, their overlap is
\begin{equation}
 \omega(A,B)=\frac{\lVert A^{\top}B\rVert_F^2}{r},
 \label{sup:eq:overlap}
\end{equation}

with random reference $r/K$; observed overlaps between dominant subspaces reach 7--12 times this reference.

\subsection{Generalization across generators and datasets}
\label{sup:sec:generalization}

Table~\ref{sup:tab:mechgen} summarizes the two-effect decomposition across generator backbones and datasets. For the cross-generator evaluations, LDM and LCM use 200 and four sampling steps, respectively, at $256\times256$ resolution, whereas SDXL uses 12 Euler sampling steps at $1024\times1024$ resolution. Across all measured settings, the interaction share $\pi_c$ remains high at 0.84--0.98 (0.95--0.98 on SDXL). The rank-8 concentration $\eta_8$ reaches 0.82--0.85 on LDM and LCM and 0.70 on RAF-DB under $E_0$, while dropping to 0.27--0.29 on SDXL. At matched selection intensity and signal correlation, main-paper Eq.~(8) associates a smaller within-source standard deviation $\sigma_v$ with a smaller expected absolute selection gain; SDXL has the lowest $\sigma_v$ among the evaluated backbones ($0.126$; Table~\ref{sup:tab:sigma}). Moreover, the leading eight interaction directions capture only $27\%$--$29\%$ of SDXL's interaction energy, indicating noise preferences spread across more directions; this more diffuse structure is consistent with the compact selector's lower gain recovery on this backbone ($53.4\%$ in main-paper Table~3).

\begin{table}[t]
\centering\small
\setlength{\tabcolsep}{4pt}
\renewcommand{\arraystretch}{1.10}

\caption{\textbf{Mechanism across settings.} Interaction share $\pi_c$ and rank-8 concentration $\eta_8$ per encoder, using 512 test sources for each of LDM, LCM, and SDXL, and 1,150 training-pool sources for RAF-DB ($E_0$ only); the default row uses the response matrices analyzed in Fig.~\ref{fig:supp_structure}.}
\label{sup:tab:mechgen}
\begin{tabular}{@{}lcc@{}}
\toprule
Setting & $\pi_c$ ($E_0$/$E_A$/$E_B$) & $\eta_8$ ($E_0$/$E_A$/$E_B$) \\
\midrule
MeanFlow B/2 (default) & 0.85 / 0.84 / 0.85 & 0.73 / 0.71 / 0.68 \\
LDM & 0.88 / 0.88 / 0.89 & 0.84 / 0.83 / 0.83 \\
LCM & 0.85 / 0.86 / 0.89 & 0.84 / 0.85 / 0.82 \\
SDXL & 0.95 / 0.98 / 0.98 & 0.29 / 0.28 / 0.27 \\
RAF-DB ($E_0$) & 0.85 & 0.70 \\
\bottomrule
\end{tabular}
\end{table}

Table~\ref{tab:v18-raf-mech} complements the RAF-DB decomposition with held-out encoders $E_A$ and $E_B$ on the 1,150-source RAF-DB training-pool response matrix. All three encoders exhibit consistent high interaction shares ($\pi_c=0.8548$--$0.9104$) and rank-8 concentrations ($\eta_8=0.6918$--$0.7142$), with rank-8 subspace overlaps of 0.346 ($E_0$--$E_A$), 0.311 ($E_0$--$E_B$), and 0.622 ($E_A$--$E_B$)---well above the random reference $8/128=0.0625$---showing that the dominant low-rank interaction structure shared across encoders is also present on RAF-DB.

\begin{table}[t]
\centering\small
\setlength{\tabcolsep}{4pt}
\renewcommand{\arraystretch}{1.11}

\caption{\textbf{Cross-encoder decomposition on RAF-DB.} Interaction share $\pi_c$, rank-8 concentration $\eta_8$, and within-source variation $\sigma_v$ on the 1,150-source RAF-DB training-pool response matrix.}
\label{tab:v18-raf-mech}
\begin{tabular}{@{}lrrr@{}}
\toprule
Encoder & $\pi_c$ & $\eta_8$ & $\sigma_v$ \\
\midrule
$E_{0}$ & 0.8548 & 0.7009 & 0.2656 \\
$E_{A}$ & 0.9017 & 0.7142 & 0.2442 \\
$E_{B}$ & 0.9104 & 0.6918 & 0.2674 \\
\bottomrule
\end{tabular}
\end{table}

\subsection{Selector architecture comparison}

\begin{figure*}[t]
  \centering
  \includegraphics[width=\textwidth]{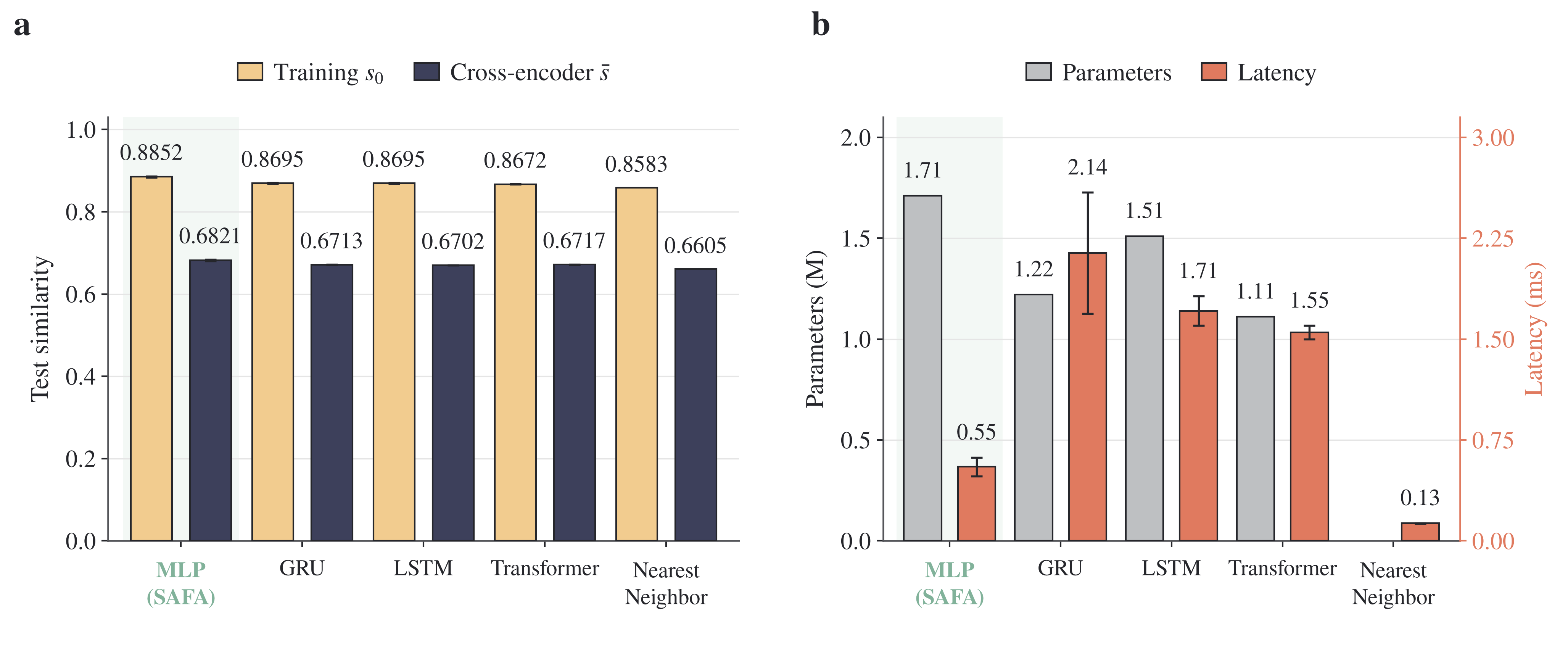}

  \caption{\textbf{Selector-architecture comparison.} (a) Test similarity. (b) Parameter count and inference latency. The highlighted \safa{} MLP corresponds to the default configuration in main-paper Table~1.}
  \label{fig:supp_selector}
\end{figure*}

Figure~\ref{fig:supp_selector} compares five selector architectures: MLP, GRU, LSTM, Transformer, and nearest-neighbor retrieval. Panel (a) reports test cosine similarity under the training encoder ($s_0$) and mean cross-encoder similarity ($\bar{s}$); panel (b) reports parameter count (millions) and inference latency (ms). Chosen on the development set, the three-layer MLP used by \safa{} has 1.71 million parameters and an inference latency of $0.55 \pm 0.07$ ms, achieving both the highest test similarity ($s_0=0.8852$, $\bar{s}=0.6821$) and the lowest latency among parametric models. All four parametric predictors outperform nearest-neighbor retrieval, with the MLP combining the highest test similarity and the lowest inference latency among them, supporting its use as a compact selector.

\section{Offline and Online Computational Cost}
\label{sup:costs}

\paragraph{Cost accounting.}

We report generator conditioning, response-bank construction, and quality labeling separately. One device-hour corresponds to one allocated GPU for one hour (e.g., four GPUs running for one hour count as four device-hours), and each stage interval includes loading, evaluation, I/O, and finalization.

\subsection{MeanFlow B/2 preparation}

Table~\ref{sup:tab:b2cost} details the offline preparation cost from main-paper Table~1 and Sec.~4.2: 7.7~h for generator conditioning, 2.7~h for candidate generation and response computation across $1{,}152\times128=147{,}456$ images, 5.2~h for quality assessment, and $\approx 18$~s for selector training, totaling 15.6~h. All noise-selection methods share the 7.7~h conditioning stage, yielding total preparation costs of 8.0~h for Good Seed and 7.9~h for NoiseQuery. For SAFA, quality assessment is performed offline; each online request runs the selector once and generates eight images without invoking any quality model.

\begin{table}[t]
\centering\small
\setlength{\tabcolsep}{4pt}
\renewcommand{\arraystretch}{1.10}

\caption{Offline preparation breakdown for MeanFlow B/2, matching main-paper Table~1 (total) and Sec.~4.2 (stages). GPU-stage costs are in device-hours and selector training is in seconds.}
\label{sup:tab:b2cost}
\begin{tabular}{@{}llr@{}}
\toprule
Stage & Workload & Cost \\
\midrule
Conditioning & \shortstack[l]{20,000 updates\\(batch 256)} & 7.7~h \\
\shortstack[l]{Response bank and\\final-PNG $E_0$} & $1{,}152\times128$ images & 2.7~h \\
Quality assessment & 147,456 images & 5.2~h \\
Selector training & One selector & $\approx 18$~s \\
\midrule
Total preparation & Main-paper Table~1 & 15.6~h \\
\bottomrule
\end{tabular}
\end{table}

\subsection{Generator optimization methods}

In main-paper Table~1, the offline costs for joint training from $G_{\text{base}}$ are reported as estimates for a unified 20,000-update budget matching the conditioning stage: 33.4~h for PCGrad, 33.0~h for CAGrad, 37.9~h for FAMO, 24.1~h for DRaFT, 19.4~h for AlignProp, and 37.6~h for LoRA. Table~\ref{sup:tab:warmcost} separately reports the measured device-hours of the full 20,000-update runs warm-started from $G_0$.

\subsection{Generator optimization warm-started from the conditioned generator}
\label{sup:sec:warmstart}

To evaluate generator optimization from a shared conditioned starting point, we re-run all six baselines from the already-conditioned $G_0$, rather than jointly from $G_{\text{base}}$ as in main-paper Table~1. Every method inherits the conditioning stage, runs for 20{,}000 updates, and is evaluated on the same 512 test sources with eight outputs per source under the same automatic ArtifactReward quality gate as the main paper; \safa{} retains its main-paper configuration ($s_0=0.885$, $\bar{s}=0.682$, $q=87.40\%$ in main-paper Tables~1 and~4). Unlike the joint runs from $G_{\text{base}}$ (which select the checkpoint with the highest development-set $s_0$), all six warm-started optimization methods are evaluated using the EMA weights of the final checkpoint after 20{,}000 updates from $G_0$, without checkpoint reselection based on development-set similarity. Starting from $G_0$ mitigates early training instability for some baselines relative to joint training from $G_{\text{base}}$ (for instance, LoRA's quality pass rate rises from 34.72\% in main-paper Table~1 to 79.69\% when warm-started from $G_0$), enabling a direct comparison from the same conditioned generator.

\begin{table}[t]
\centering\small
\setlength{\tabcolsep}{3pt}
\renewcommand{\arraystretch}{1.10}

\caption{Warm-started batch from $G_0$ (512 test sources, eight outputs per source). The six generator optimization baselines use the final EMA checkpoint after 20{,}000 updates from $G_0$, while SAFA retains its main-paper configuration as reference. $\bar{s}$ is the mean cross-encoder similarity across Swin-T and MobileNetV3; $D_8$ is the maximum size of a mutually distinct subset per source (pairwise LPIPS $\geq\delta$) without quality or similarity gating; $U_8$ is the maximum subset size per source jointly satisfying quality, similarity ($s_0\geq\tau$), and pairwise LPIPS $\geq\delta$. Both $D_8$ and $U_8$ are averaged equally over test sources.}
\label{sup:tab:warmmetrics}
\begin{tabular}{@{}lcccccc@{}}
\toprule
Method & $s_0$ & $\bar{s}$ & $q$ (\%) & $D_8$ & $U_8$ & \shortstack{Pairwise\\LPIPS} \\
\midrule
\safa{} & 0.885 & 0.682 & 87.40 & 8.000 & 5.85 & 0.452 \\
LoRA & 0.888 & 0.684 & 79.69 & 8.000 & 5.17 & 0.387 \\
CAGrad & 0.991 & 0.793 & 39.67 & 1.000 & 0.83 & 0.010 \\
PCGrad & 0.991 & 0.775 & 43.99 & 1.000 & 0.98 & 0.015 \\
DRaFT & 0.990 & 0.796 & 15.06 & 1.000 & 0.45 & 0.023 \\
FAMO & 0.992 & 0.768 & 43.31 & 1.000 & 0.78 & 0.006 \\
AlignProp & 0.448 & 0.223 & 12.01 & 1.002 & 0.02 & 0.009 \\
\bottomrule
\end{tabular}
\end{table}

As shown in Table~\ref{sup:tab:warmmetrics}, even when warm-started from $G_0$ for the full 20{,}000-update budget, none of the six generator-optimization baselines matches \safa{} across similarity, diversity, and image quality simultaneously. CAGrad, PCGrad, DRaFT, and FAMO push $s_0\geq0.99$ while collapsing across outputs ($D_8=1.000$, mean pairwise LPIPS $\leq0.023$), yielding at most one distinct image per source alongside severely degraded quality pass rates (15.06\%--43.99\% vs.\ 87.40\% for \safa{}) and low usable yields ($U_8=0.45\text{--}0.98$ vs.\ $5.85$). AlignProp degrades across similarity ($s_0=0.448$, $\bar{s}=0.223$), diversity ($D_8=1.002$, mean pairwise LPIPS $0.009$), quality ($q=12.01\%$), and usable yield ($U_8=0.02$). Among all six baselines, only warm-started LoRA preserves output-set diversity ($D_8=8.000$, albeit with lower mean pairwise LPIPS than \safa{}, $0.387$ vs.\ $0.452$) and achieves slightly higher mean similarities ($s_0=0.888$ vs.\ $0.885$; $\bar{s}=0.684$ vs.\ $0.682$), while trailing \safa{} by 7.71 percentage points in quality pass rate (79.69\% vs.\ 87.40\%) and by 0.68 images per source in usable yield ($U_8=5.17$ vs.\ $5.85$). Moreover, across all six warm-started runs, generator optimization incurs higher total offline cost (17.679--41.029~device-hours in Table~\ref{sup:tab:warmcost}) than \safa{}'s complete preparation cost (15.6~device-hours).

Overall, even when warm-started from $G_0$ with higher offline compute, none of the six generator-optimization baselines matches \safa{}'s joint balance of similarity, diversity, image quality, and usable yield.

\begin{table}[t]
\centering\small
\setlength{\tabcolsep}{6pt}
\renewcommand{\arraystretch}{1.10}

\caption{Measured offline costs (device-hours) for the warm-started 20{,}000-update runs. ``With shared $G_0$'' includes the 7.7~h conditioning stage; \safa{} reflects the main-paper preparation cost.}
\label{sup:tab:warmcost}
\begin{tabular}{@{}lrr@{}}
\toprule
Method & Beyond $G_0$ & With shared $G_0$ \\
\midrule
\safa{} (main-paper reference) & 7.9 & 15.6 \\
LoRA & 12.379 & 20.057 \\
CAGrad & 14.386 & 22.063 \\
PCGrad & 10.625 & 18.303 \\
DRaFT & 10.002 & 17.679 \\
FAMO & 33.352 & 41.029 \\
AlignProp & 12.155 & 19.832 \\
\bottomrule
\end{tabular}
\end{table}

\subsection{Conditioning across generator backbones}

Conditioning or distillation costs, in device-hours, are 7.7~h for MeanFlow B/2, 3.238~h for B/4, 26.379~h for L/2, 18.444~h for LDM, and 14.398~h for LCM. The B/2 entry is the conditioning stage already listed in Table~\ref{sup:tab:b2cost}. The LCM entry covers distillation only; conditioning of its teacher is accounted for under LDM.

\paragraph{Online generation.}

At inference, \safa{} evaluates its selector once and selects eight dictionary indices; each source request returns eight generated images. The offline quality model and response-bank construction do not enter this request. Best-of-$N$ generates and evaluates the candidate set before selection. Generator conditioning and bank construction are amortized over subsequent augmentation requests.

\paragraph{Amortization and break-even.}

Under the device-hour accounting reported in main-paper Table~1, \safa{} adds $15.6-7.7=7.9$~device-hours of one-time preparation relative to Best-of-$N$ sharing $G_0$. For an added offline cost $\Delta T$ in device-hours and a per-request saving $\Delta t$ in seconds, the break-even number of source requests is
\begin{equation}
n_{\text{break}}=\bigl\lceil 3600\,\Delta T/\Delta t\bigr\rceil.
\label{sup:eq:breakeven}
\end{equation}

Using the rounded timings in main-paper Table~1 and Sec.~4.5 ($20.4$~s), \safa{} saves $20.4-0.15=20.25$~s per source request relative to quality-penalized Best-of-$N$, yielding a break-even point of $\lceil7.9\times3600/20.25\rceil=1{,}405$ source requests (consistent with ``about 1,400'' in the main paper). Against unpenalized Best-of-$N$, the per-request saving is $2.35-0.15=2.20$~s, giving $12{,}928$ source requests under the same formula. These break-even counts follow directly from those rounded main-paper timings.

\paragraph{Quality supervision trade-off.}

Quality assessment accounts for about one third of \safa{}'s preparation cost ($5.2/15.6$). Main-paper Table~4 reports that increasing the penalty weight $w$ from $0$ to $0.2$ raises the quality pass rate by 20.6 percentage points and improves usable yield $U_8$ from $4.71$ to $5.85$, with cross-encoder similarity $\bar{s}=0.682$ at the default setting. Supplementary single-seed evaluations give $U_8=5.60$, $5.82$, and $5.80$ at $w=0.05$, $0.1$, and $0.4$, respectively, with $\bar{s}$ ranging from $0.6719$ to $0.6821$ across the five settings (Table~\ref{sup:tab:wsens}). At $w=0.4$, the pass rate reaches 89.6\% while usable yield is $5.80$; the default $w=0.2$ combines an 87.4\% pass rate with $U_8=5.85$, offering a favorable balance between quality and usable yield.

Because the penalty targets artifact suppression, higher baseline pass rates leave less headroom for further quality gains.

\begin{table}[t]
\centering\small
\setlength{\tabcolsep}{5pt}
\renewcommand{\arraystretch}{1.10}

\caption{Sensitivity to the quality penalty weight $w$. Rows for $w=0$ and $w=0.2$ reproduce main-paper Table~4; $w=0.05,0.1,0.4$ are supplementary single-seed selector evaluations.}
\label{sup:tab:wsens}
\begin{tabular}{@{}lcccc@{}}
\toprule
$w$ & $q$ (\%) & $U_8$ & $\bar{s}$ & $s_0$ \\
\midrule
0 & 66.8 & 4.71 & 0.6765 & 0.8820 \\
0.05 & 79.5 & 5.60 & 0.6790 & 0.8803 \\
0.1 & 84.5 & 5.82 & 0.6790 & 0.8751 \\
0.2 & 87.4 & 5.85 & 0.6821 & 0.8852 \\
0.4 & 89.6 & 5.80 & 0.6719 & 0.8637 \\
\bottomrule
\end{tabular}
\end{table}

\paragraph{Within-source variation across backbones.}

Equation~(8) of the main paper models the expected selection gain through selection intensity, signal correlation, and within-source variation. SDXL exhibits smaller within-source variation ($\sigma_v=0.126$) than MeanFlow B/2 ($0.223$; Table~\ref{sup:tab:sigma}), consistent with its smaller absolute selection gain in main-paper Table~3. Even in this narrower selection regime, \safa{} improves cross-encoder similarity on SDXL from $0.735$ under random sampling to $0.776$ with only eight generations (an absolute gain of $0.041$), demonstrating the benefit of source-specific noise selection on this backbone.

\begin{table}[t]
\centering\small
\setlength{\tabcolsep}{6pt}
\renewcommand{\arraystretch}{1.10}

\caption{Within-source variation $\sigma_v$ across generator backbones. For each source, we compute the population standard deviation of $v=(s_{E_A}+s_{E_B})/2$ over all $K=128$ candidates, then average equally over the 512 test sources. All candidates are included.}
\label{sup:tab:sigma}
\begin{tabular}{@{}lc@{}}
\toprule
Generator & $\sigma_v$ \\
\midrule
MeanFlow B/2 & 0.223 \\
LDM & 0.252 \\
LCM & 0.213 \\
SDXL & 0.126 \\
\bottomrule
\end{tabular}
\end{table}

\section{Recognition with Scarce Expression Labels}
\label{sup:sec:downstream}

The main-paper recognition setting uses 1{,}200 labeled AffectNet-S training images and evaluates on the official 3{,}969-image validation set. In main-paper Table~2, \safa{} achieves $39.7\pm0.4\%$ on ResNet-18 and $43.3\pm1.2\%$ on ConvNeXt-T over five seeds, representing relative gains of $16.8\%$ and $17.7\%$ over real images alone ($34.0\pm0.6\%$ and $36.8\pm1.3\%$). Here, we supplement these results with cross-dataset transfer to RAF-DB and per-class recall on scarce expressions.

\subsection{Cross-dataset transfer and scarce-class recall}
\label{sup:sec:downstream-extra}

To assess cross-dataset generalization, we evaluate classifiers trained on AffectNet-S and checkpoint-selected using AffectNet development data directly on the 3{,}071-image RAF-DB test set. Predictions are obtained by taking the argmax over the seven shared AffectNet output classes in their original logit order, followed by mapping to RAF-DB labels. Table~\ref{sup:tab:raf-transfer} shows higher mean transfer accuracy on both backbones over five classifier seeds: from $25.39\pm2.67\%$ to $27.84\pm4.17\%$ on ResNet-18 and from $29.24\pm3.13\%$ to $35.11\pm3.66\%$ on ConvNeXt-T, corresponding to gains of $2.45$ and $5.87$ percentage points.

\begin{table}[t]
\centering
\small
\setlength{\tabcolsep}{6pt}
\renewcommand{\arraystretch}{1.10}

\caption{Cross-dataset AffectNet-S to RAF-DB recognition accuracy
(\%, mean $\pm$ sample SD over five classifier seeds).
Classifiers are trained on AffectNet-S and checkpoint-selected
using AffectNet development data.}
\label{sup:tab:raf-transfer}
\begin{tabular}{@{}lrr@{}}
\toprule
Training arm & ResNet-18 & ConvNeXt-T \\
\midrule
Real-only & $25.39\pm2.67$ & $29.24\pm3.13$ \\
+\safa{} & $27.84\pm4.17$ & $35.11\pm3.66$ \\
\bottomrule
\end{tabular}
\end{table}

Table~\ref{sup:tab:fer-class} reports recall on the official AffectNet validation set for the three classes with the fewest real training images in AffectNet-S---Happiness, Sadness, and Surprise, each with only 50 real training images---in the supplementary recognition experiments (Sec.~\ref{sup:sec:classifier-controls}). Relative to real images alone, +\safa{} improves mean recall across all three 50-image classes on both backbones: Surprise recall rises from $10.39\%$ to $24.46\%$ on ResNet-18 and from $11.08\%$ to $31.81\%$ on ConvNeXt-T, Happiness recall rises from $38.01\%$ to $53.31\%$ and from $46.25\%$ to $56.15\%$, and Sadness recall rises from $5.19\%$ to $8.45\%$ and from $9.26\%$ to $14.97\%$, respectively.

Across the six backbone--class comparisons, the offline Best-of-$N$ reference achieves a further $1.78\text{--}5.23$ percentage points of mean recall, with the largest gap on Sadness using ConvNeXt-T ($14.97\%$ vs.\ $20.20\%$). However, Best-of-$N$ generates and ranks the full $K=128$ candidate pool per training source and inspects up to the top 32 ranked candidates to fill the same per-source augmentation quota (Sec.~\ref{sup:sec:classifier-controls}). By contrast, \safa{} selects noise vectors before generation without generating the full candidate pool, while exhibiting smaller observed cross-seed standard deviations than Best-of-$N$ across all three classes on ConvNeXt-T ($2.99$ vs.\ $12.85$ percentage points for Happiness, $6.03$ vs.\ $8.35$ for Sadness, and $7.07$ vs.\ $14.11$ for Surprise).

Across all eight classes in these supplementary runs, +\safa{} also raises macro-F1 from $31.50\pm0.70\%$ to $37.05\pm0.38\%$ on ResNet-18 and from $35.01\pm1.33\%$ to $42.15\pm0.76\%$ on ConvNeXt-T, while offline Best-of-$N$ achieves $38.02\pm1.10\%$ and $43.09\pm1.53\%$, respectively.

\begin{table*}[t]
\centering
\small
\setlength{\tabcolsep}{4pt}
\renewcommand{\arraystretch}{1.11}

\caption{Recall on the official AffectNet validation set for the three classes with the fewest real training images in AffectNet-S (50 images per class: Happiness, Sadness, and Surprise) in the supplementary recognition experiments ($\%$, mean $\pm$ sample SD over five classifier seeds). Support denotes test images per seed; BoN ranks the 128-candidate pool per training source and filters by quality and distinctness to fill the same augmentation quota (Sec.~\ref{sup:sec:classifier-controls}).}
\label{sup:tab:fer-class}
\begin{tabular}{@{}llrrrr@{}}
\toprule
Backbone & Class & Support & Real-only & +\safa{} & BoN \\
\midrule
ResNet-18 & Happiness & 493 & $38.01\pm3.43$ & $53.31\pm4.22$ & $55.86\pm6.55$ \\
ResNet-18 & Sadness & 497 & $5.19\pm1.79$ & $8.45\pm3.39$ & $11.23\pm2.93$ \\
ResNet-18 & Surprise & 493 & $10.39\pm2.92$ & $24.46\pm5.21$ & $26.82\pm7.53$ \\
ConvNeXt-T & Happiness & 493 & $46.25\pm9.38$ & $56.15\pm2.99$ & $59.15\pm12.85$ \\
ConvNeXt-T & Sadness & 497 & $9.26\pm5.44$ & $14.97\pm6.03$ & $20.20\pm8.35$ \\
ConvNeXt-T & Surprise & 493 & $11.08\pm5.58$ & $31.81\pm7.07$ & $33.59\pm14.11$ \\
\bottomrule
\end{tabular}
\end{table*}

\section{Additional Qualitative Results}

\label{sec:supp_qualitative}

\subsection{Expanded comparison across methods}

Figures~\ref{fig:supp_cases_a}--\ref{fig:supp_cases_c} compare eleven methods on four sources using MeanFlow B/2, with four displayed outputs per method and source (176 images). Each column corresponds to one source, with the same sources and column order across all three panels; each method's four outputs form a two-by-two grid. Matching local insets enlarge the mouth, eye, and brow regions, and each output displays its training-encoder cosine similarity $s_0$. The six generator optimization baselines use the checkpoints trained from $G_{\mathrm{base}}$ for main-paper Table~1, displaying outputs from fixed noise indices 0--3 for each source. In main-paper Fig.~7, the insets highlight the mouth in rows 1--2. In row 3, they highlight the brow and eyes except in the two DRaFT outputs, where they highlight the mouth.

\renewcommand{\thefigure}{S4a}

\begin{figure*}[p]
  \centering
  \includegraphics[page=1,width=\textwidth,height=0.80\textheight,keepaspectratio]{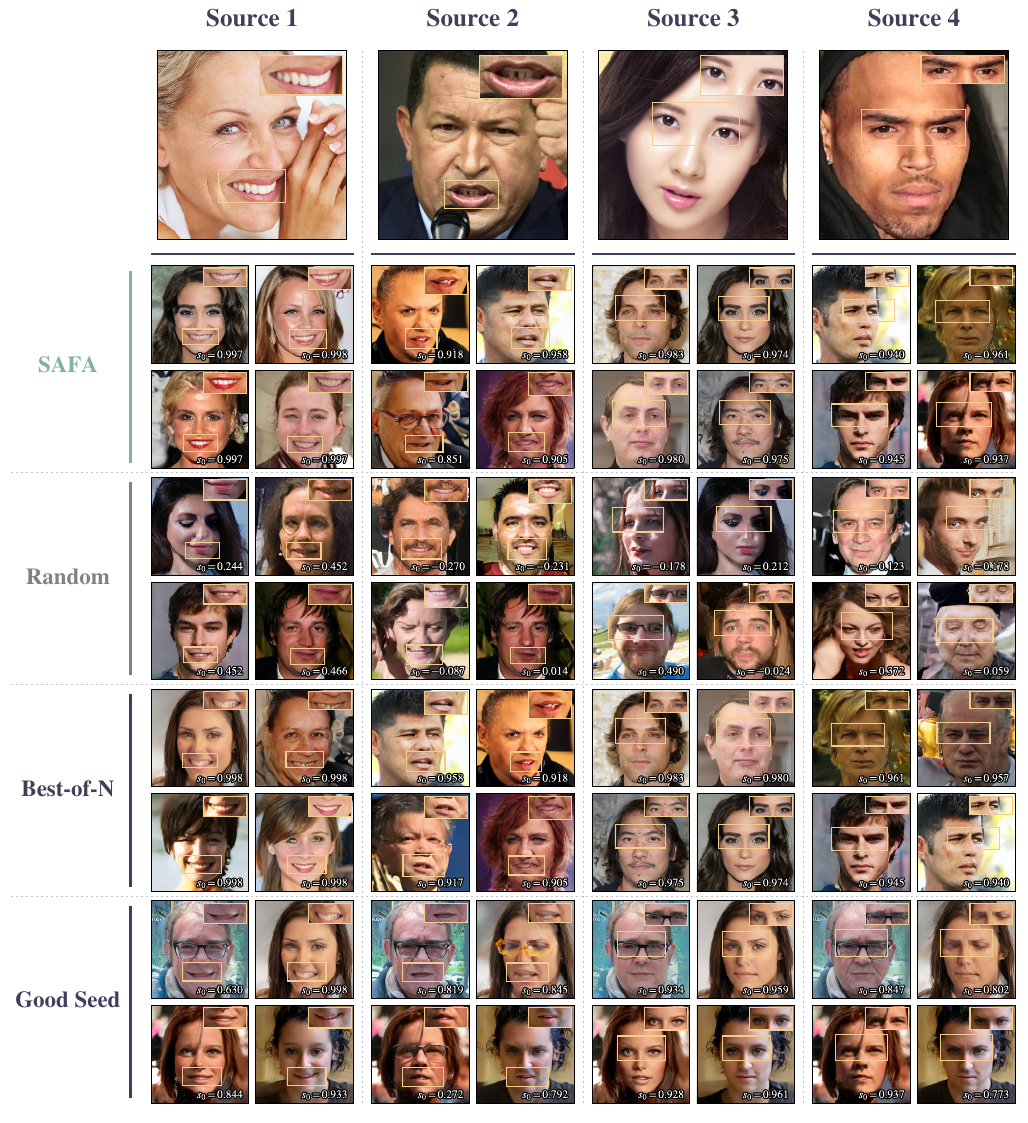}

  \caption{\textbf{Four-source qualitative comparison: SAFA, random sampling, Best-of-$N$, and Good Seed.}}
  \label{fig:supp_cases_a}
\end{figure*}

\renewcommand{\thefigure}{S4b}

\begin{figure*}[p]
  \centering
  \includegraphics[page=2,width=\textwidth,height=0.80\textheight,keepaspectratio]{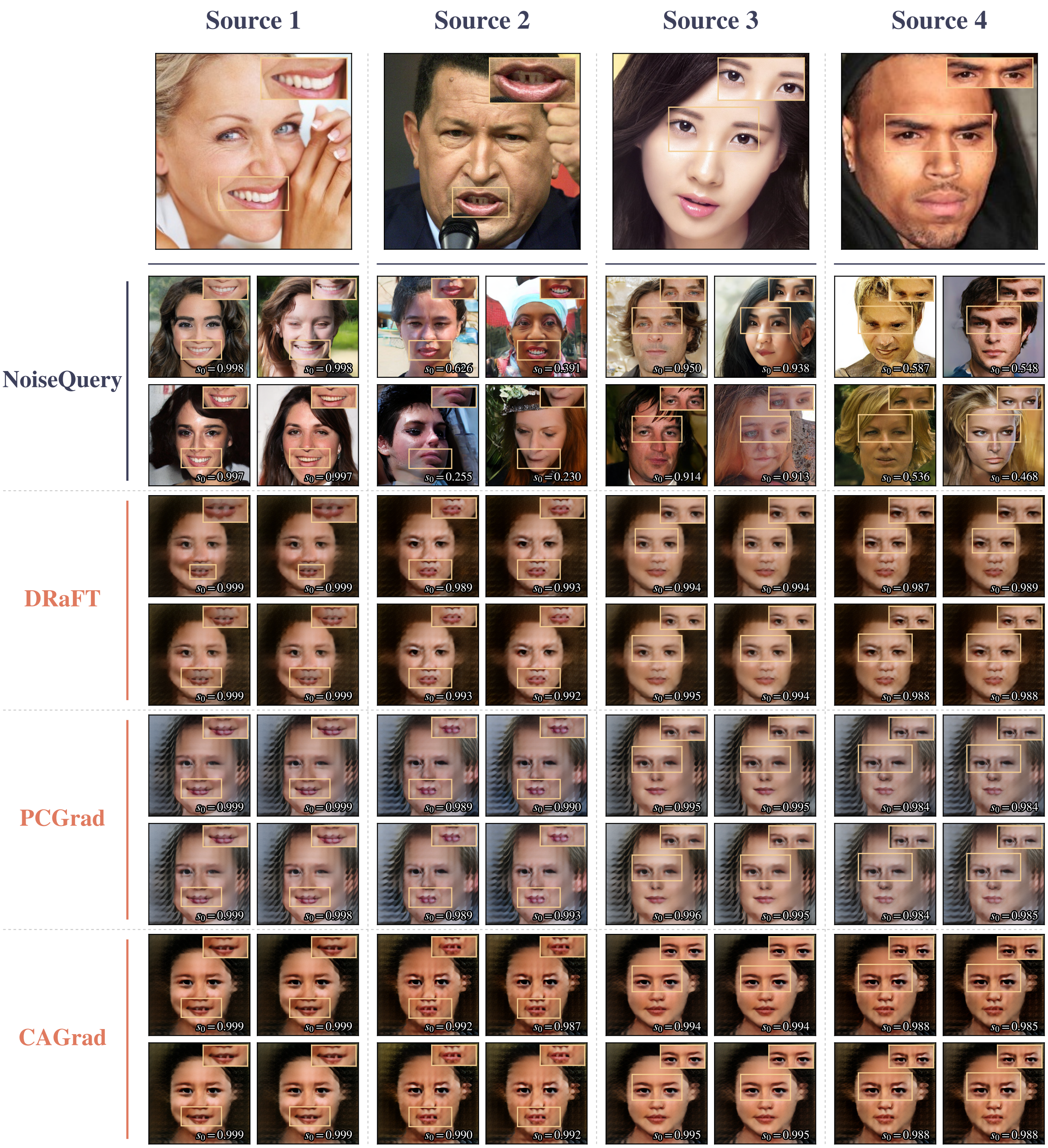}

  \caption{\textbf{Four-source qualitative comparison: NoiseQuery, DRaFT, PCGrad, and CAGrad.}}
  \label{fig:supp_cases_b}
\end{figure*}

\renewcommand{\thefigure}{S4c}

\begin{figure*}[p]
  \centering
  \includegraphics[page=3,width=\textwidth,height=0.80\textheight,keepaspectratio]{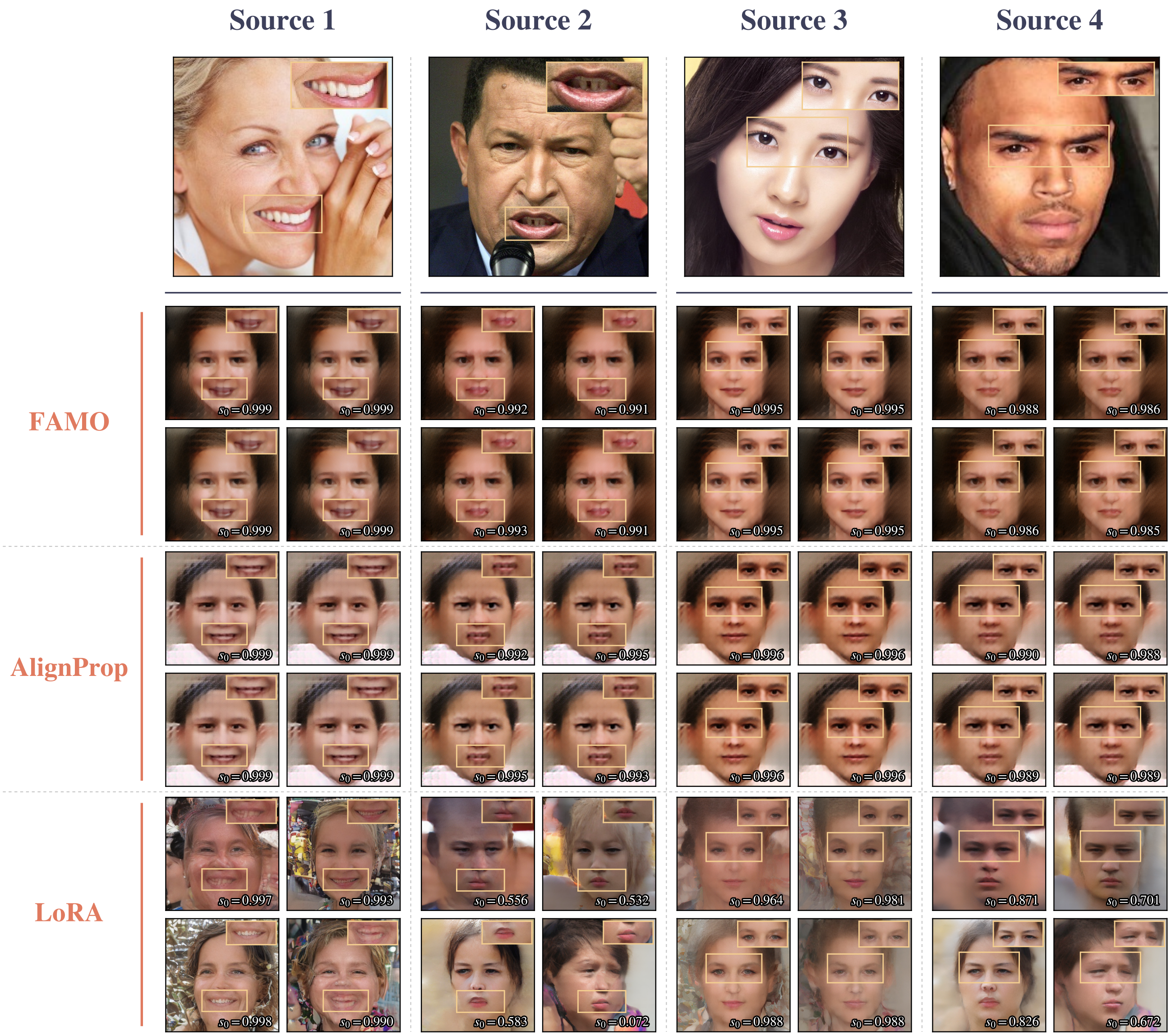}

  \caption{\textbf{Four-source qualitative comparison: FAMO, AlignProp, and LoRA.}}
  \label{fig:supp_cases_c}
\end{figure*}

\renewcommand{\thefigure}{S\arabic{figure}}
\setcounter{figure}{5}

\subsection{Additional sources}

These fifteen additional sources (Figures~\ref{fig:supp:gallery_a}--\ref{fig:supp:gallery_c}) extend the qualitative comparison of Figures~\ref{fig:supp_cases_a}--\ref{fig:supp_cases_c}, showing additional source expressions and generated appearances. Random and the four generator optimization baselines display noise index 0 across all sources, whereas \safa{} displays its two highest-ranked outputs per source in their original returned order. The optimization baselines follow the initialization setting of main-paper Table~1, training from the native pretrained backbone with a fresh conditioning adapter. In the displayed \safa{} outputs, brow and eye cues are generally preserved (e.g., b2\_0003 and b2\_0009), while mouth movements are less pronounced in some samples; for example, the wide mouth opening in source b2\_0004 is not fully retained.

\renewcommand{\thefigure}{S5a}

\begin{figure*}[p]
  \centering
  \setlength{\tabcolsep}{1pt}
  {\footnotesize
  \begin{tabular}{@{}rcccccccc@{}}
   & Source & Random & \multicolumn{2}{c}{SAFA} & PCGrad & CAGrad & FAMO & AlignProp \\
  b2\_0002 &
  \includegraphics[height=19mm]{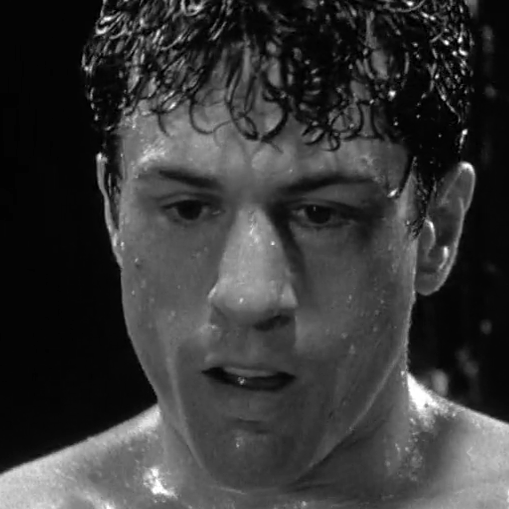} &
  \includegraphics[height=19mm]{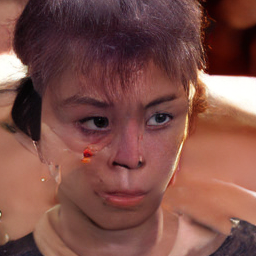} &
  \includegraphics[height=19mm]{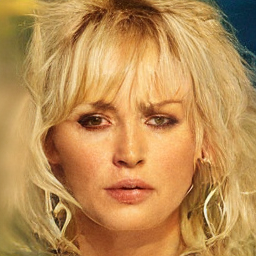} &
  \includegraphics[height=19mm]{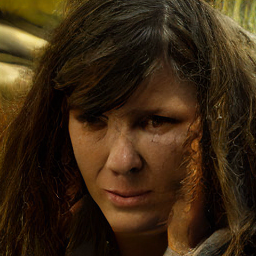} &
  \includegraphics[height=19mm]{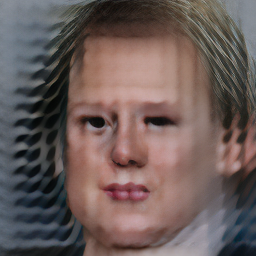} &
  \includegraphics[height=19mm]{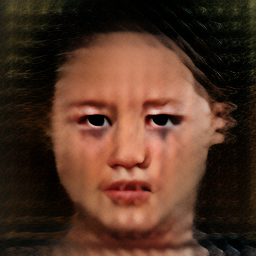} &
  \includegraphics[height=19mm]{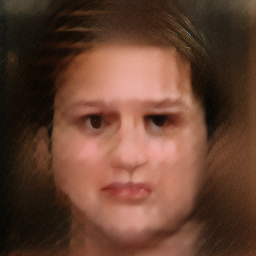} &
  \includegraphics[height=19mm]{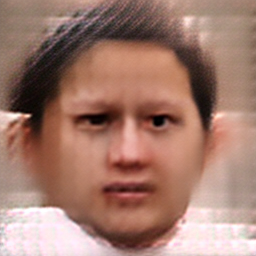} \\[2pt]
  b2\_0003 &
  \includegraphics[height=19mm]{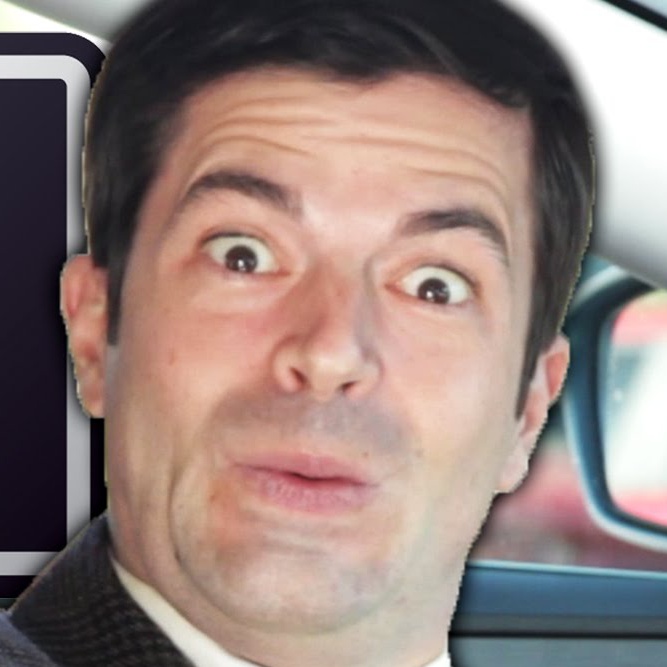} &
  \includegraphics[height=19mm]{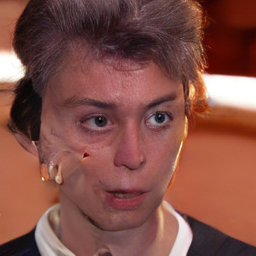} &
  \includegraphics[height=19mm]{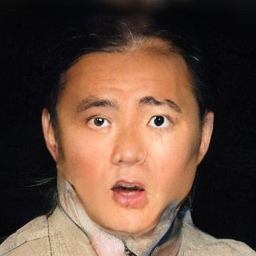} &
  \includegraphics[height=19mm]{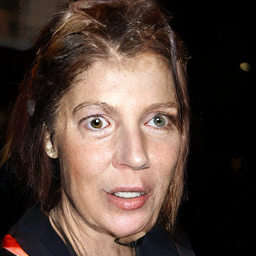} &
  \includegraphics[height=19mm]{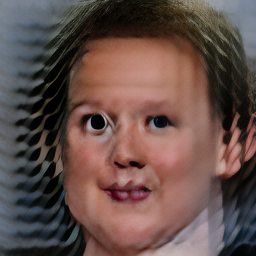} &
  \includegraphics[height=19mm]{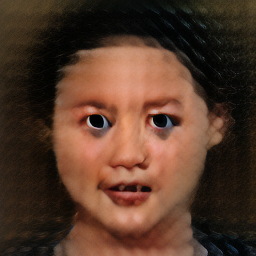} &
  \includegraphics[height=19mm]{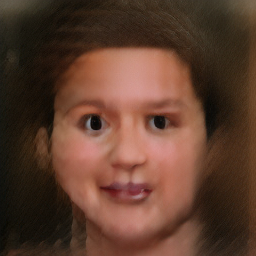} &
  \includegraphics[height=19mm]{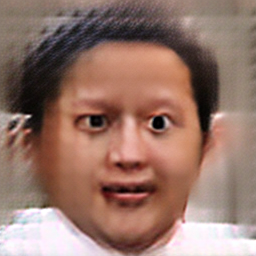} \\[2pt]
  b2\_0004 &
  \includegraphics[height=19mm]{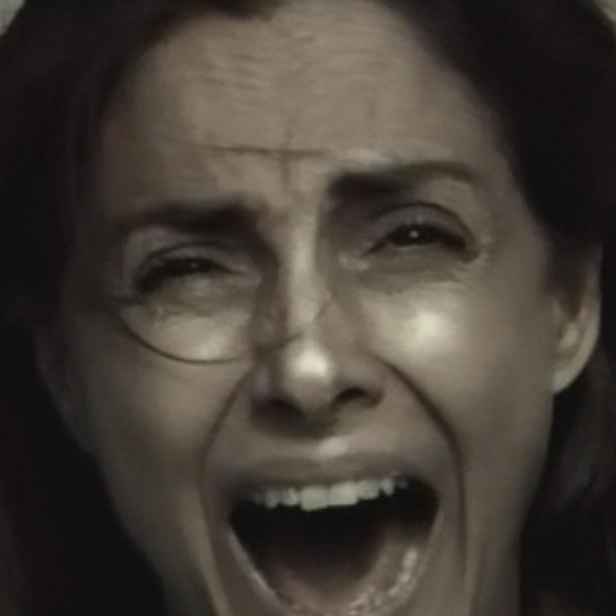} &
  \includegraphics[height=19mm]{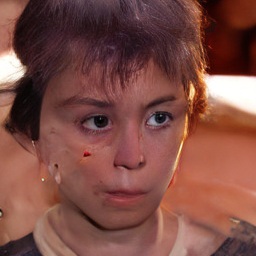} &
  \includegraphics[height=19mm]{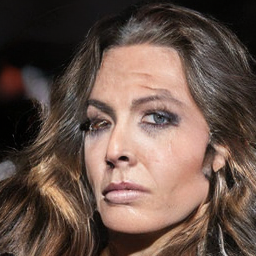} &
  \includegraphics[height=19mm]{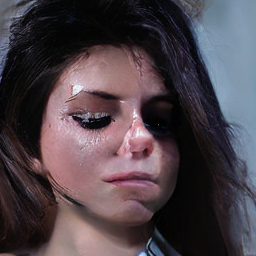} &
  \includegraphics[height=19mm]{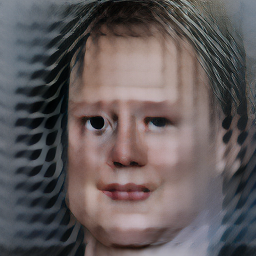} &
  \includegraphics[height=19mm]{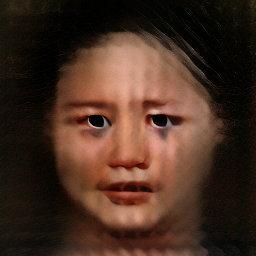} &
  \includegraphics[height=19mm]{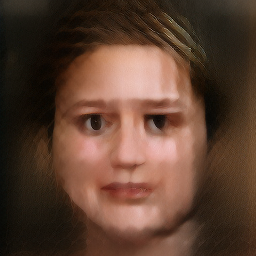} &
  \includegraphics[height=19mm]{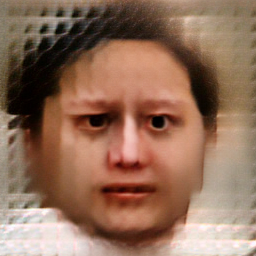} \\[2pt]
  b2\_0005 &
  \includegraphics[height=19mm]{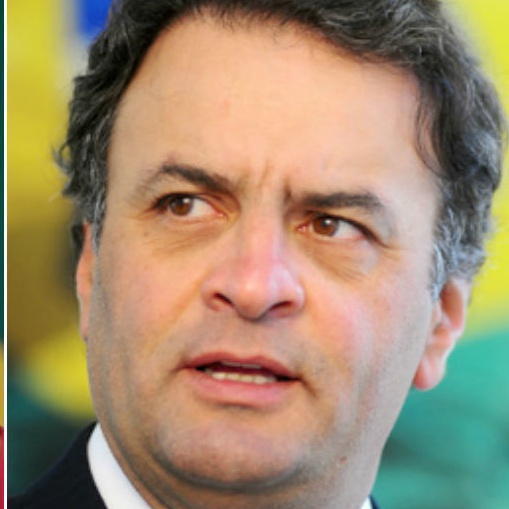} &
  \includegraphics[height=19mm]{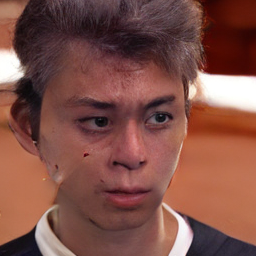} &
  \includegraphics[height=19mm]{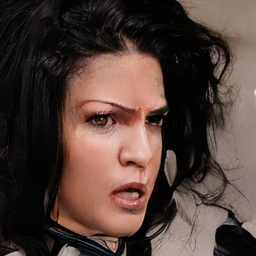} &
  \includegraphics[height=19mm]{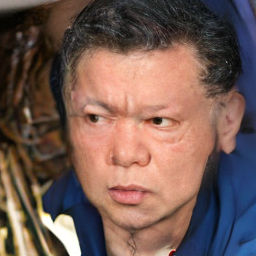} &
  \includegraphics[height=19mm]{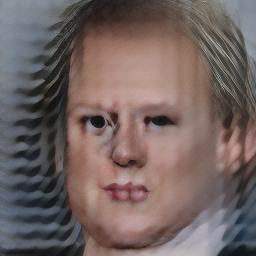} &
  \includegraphics[height=19mm]{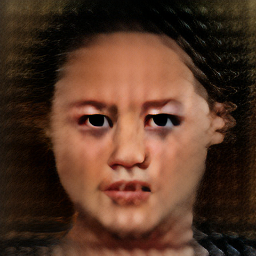} &
  \includegraphics[height=19mm]{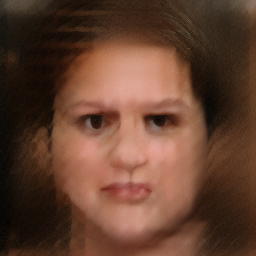} &
  \includegraphics[height=19mm]{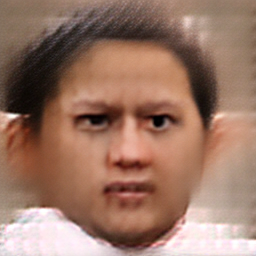} \\[2pt]
  b2\_0006 &
  \includegraphics[height=19mm]{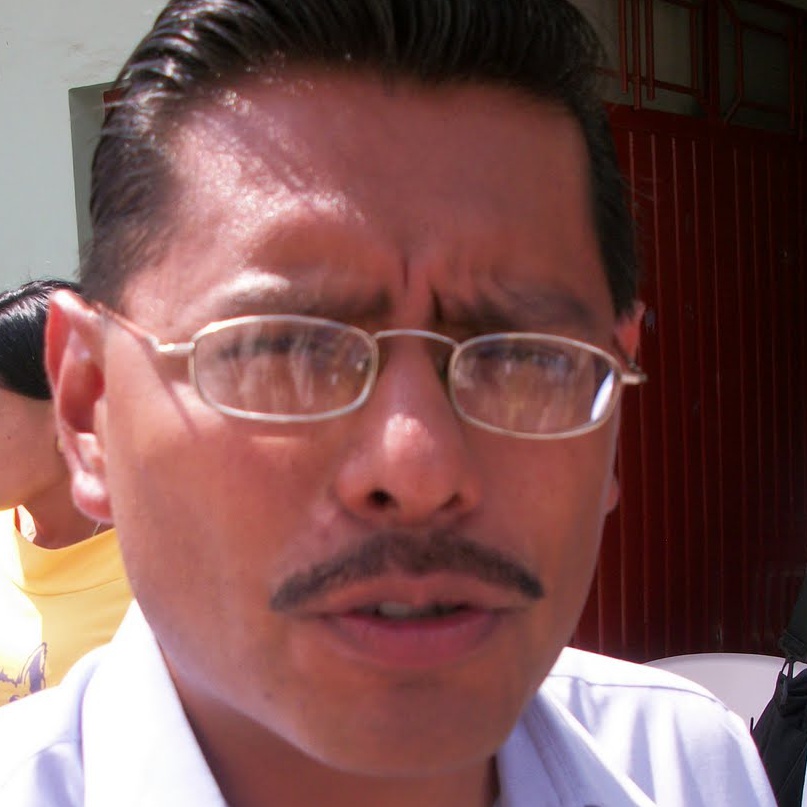} &
  \includegraphics[height=19mm]{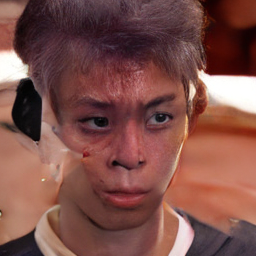} &
  \includegraphics[height=19mm]{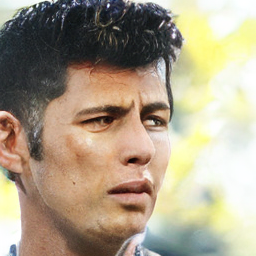} &
  \includegraphics[height=19mm]{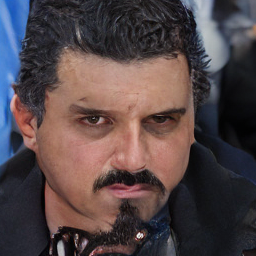} &
  \includegraphics[height=19mm]{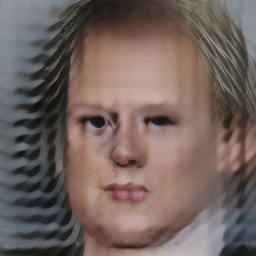} &
  \includegraphics[height=19mm]{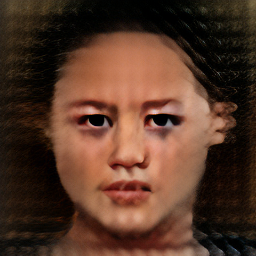} &
  \includegraphics[height=19mm]{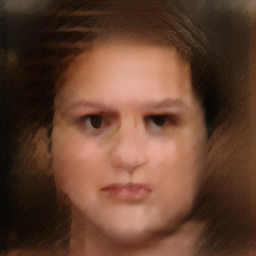} &
  \includegraphics[height=19mm]{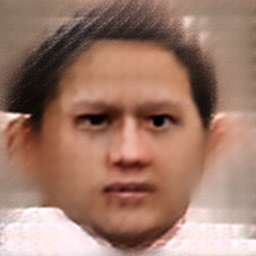} \\
  \end{tabular}}

  \caption{\textbf{Additional sources.} Rows are sources b2\_0002--b2\_0006. Columns: source image, random sampling, two SAFA outputs, PCGrad, CAGrad, FAMO, and AlignProp, all from MeanFlow B/2. Random and optimization baselines use noise index 0 throughout; SAFA shows its top two outputs per source.}
  \label{fig:supp:gallery_a}
\end{figure*}

\renewcommand{\thefigure}{S5b}

\begin{figure*}[p]
  \centering
  \setlength{\tabcolsep}{1pt}
  {\footnotesize
  \begin{tabular}{@{}rcccccccc@{}}
   & Source & Random & \multicolumn{2}{c}{SAFA} & PCGrad & CAGrad & FAMO & AlignProp \\
  b2\_0007 &
  \includegraphics[height=19mm]{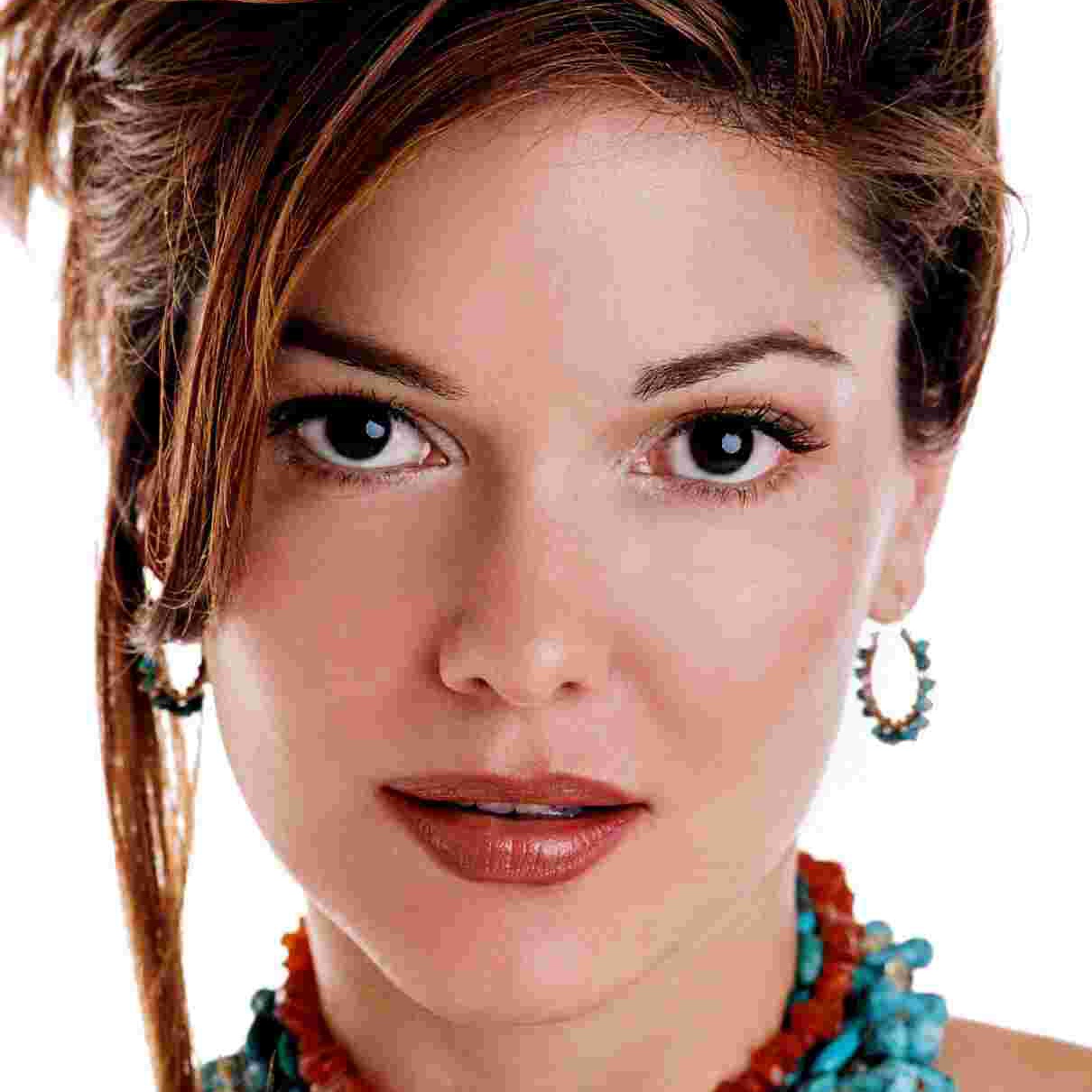} &
  \includegraphics[height=19mm]{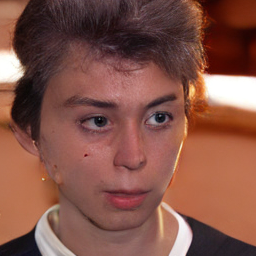} &
  \includegraphics[height=19mm]{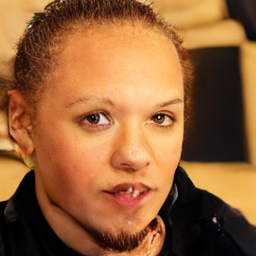} &
  \includegraphics[height=19mm]{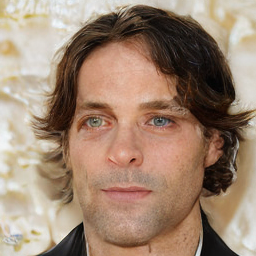} &
  \includegraphics[height=19mm]{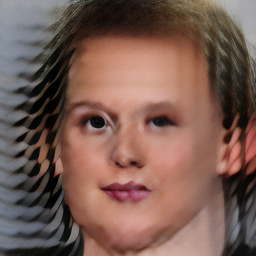} &
  \includegraphics[height=19mm]{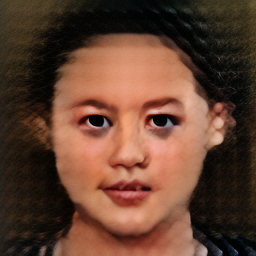} &
  \includegraphics[height=19mm]{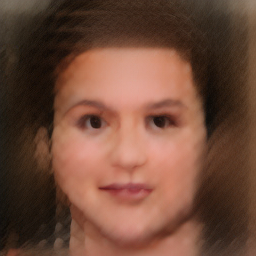} &
  \includegraphics[height=19mm]{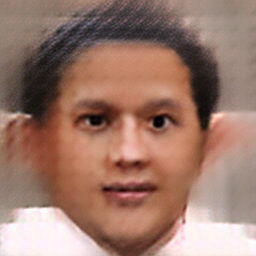} \\[2pt]
  b2\_0008 &
  \includegraphics[height=19mm]{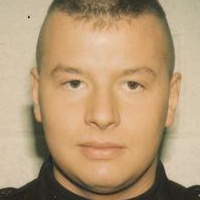} &
  \includegraphics[height=19mm]{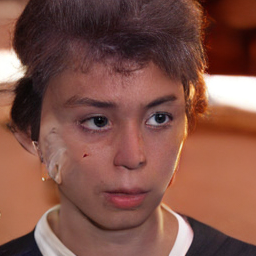} &
  \includegraphics[height=19mm]{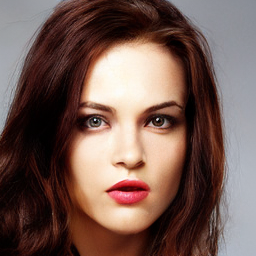} &
  \includegraphics[height=19mm]{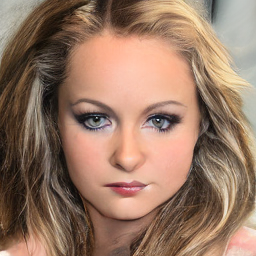} &
  \includegraphics[height=19mm]{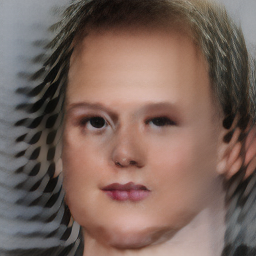} &
  \includegraphics[height=19mm]{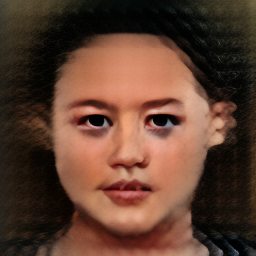} &
  \includegraphics[height=19mm]{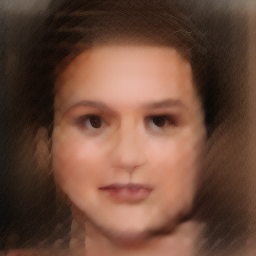} &
  \includegraphics[height=19mm]{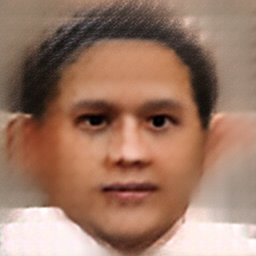} \\[2pt]
  b2\_0009 &
  \includegraphics[height=19mm]{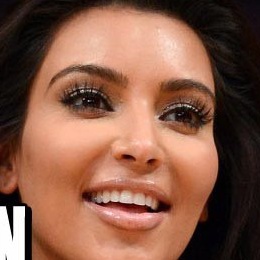} &
  \includegraphics[height=19mm]{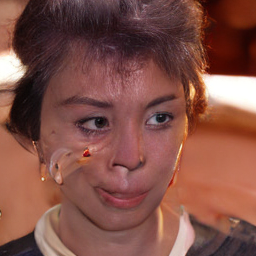} &
  \includegraphics[height=19mm]{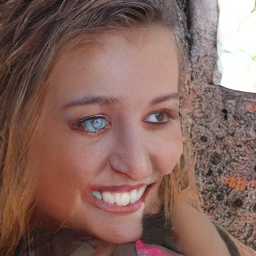} &
  \includegraphics[height=19mm]{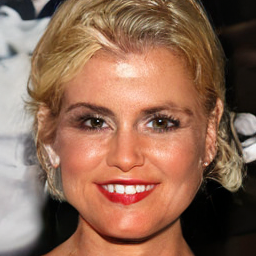} &
  \includegraphics[height=19mm]{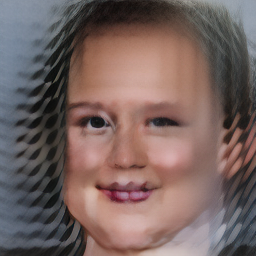} &
  \includegraphics[height=19mm]{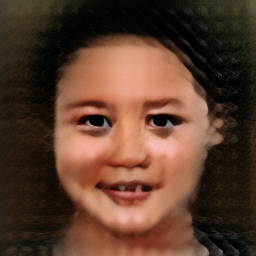} &
  \includegraphics[height=19mm]{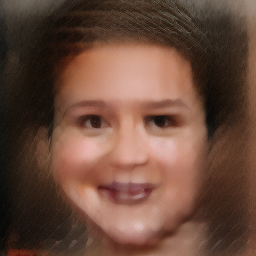} &
  \includegraphics[height=19mm]{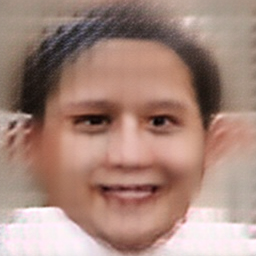} \\[2pt]
  b2\_0010 &
  \includegraphics[height=19mm]{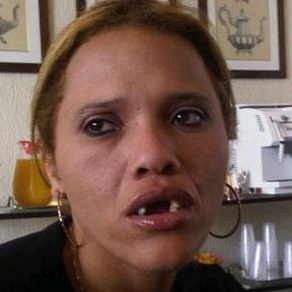} &
  \includegraphics[height=19mm]{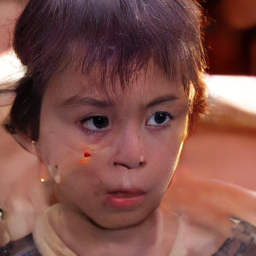} &
  \includegraphics[height=19mm]{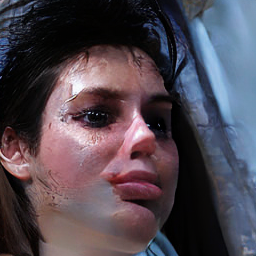} &
  \includegraphics[height=19mm]{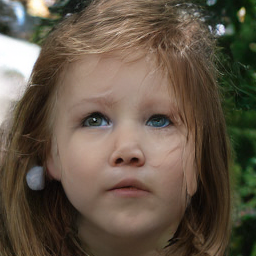} &
  \includegraphics[height=19mm]{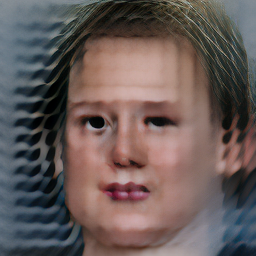} &
  \includegraphics[height=19mm]{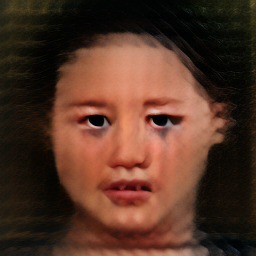} &
  \includegraphics[height=19mm]{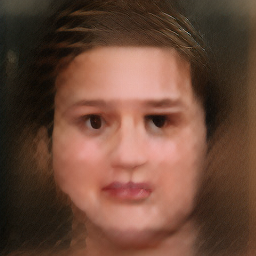} &
  \includegraphics[height=19mm]{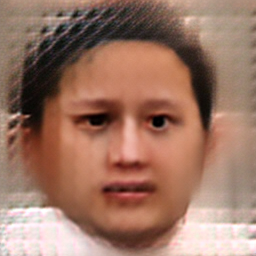} \\[2pt]
  b2\_0011 &
  \includegraphics[height=19mm]{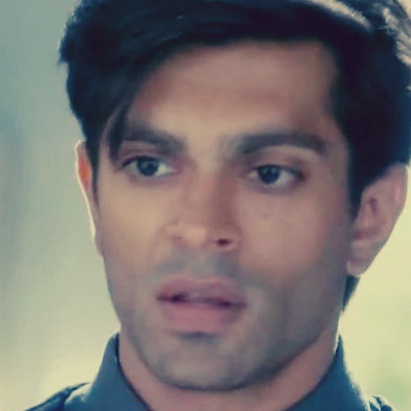} &
  \includegraphics[height=19mm]{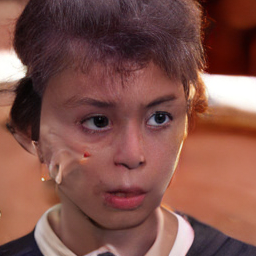} &
  \includegraphics[height=19mm]{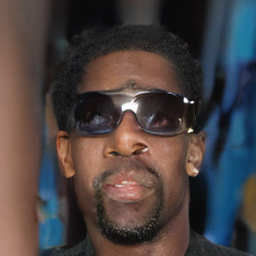} &
  \includegraphics[height=19mm]{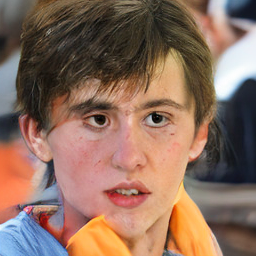} &
  \includegraphics[height=19mm]{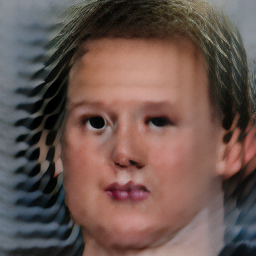} &
  \includegraphics[height=19mm]{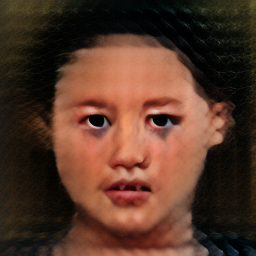} &
  \includegraphics[height=19mm]{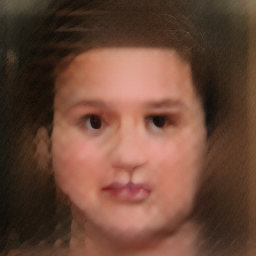} &
  \includegraphics[height=19mm]{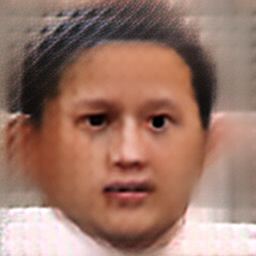} \\
  \end{tabular}}

  \caption{\textbf{Additional sources.} Rows are sources b2\_0007--b2\_0011. Columns: source image, random sampling, two SAFA outputs, PCGrad, CAGrad, FAMO, and AlignProp, all from MeanFlow B/2. Random and optimization baselines use noise index 0 throughout; SAFA shows its top two outputs per source.}
  \label{fig:supp:gallery_b}
\end{figure*}

\renewcommand{\thefigure}{S5c}

\begin{figure*}[p]
  \centering
  \setlength{\tabcolsep}{1pt}
  {\footnotesize
  \begin{tabular}{@{}rcccccccc@{}}
   & Source & Random & \multicolumn{2}{c}{SAFA} & PCGrad & CAGrad & FAMO & AlignProp \\
  b2\_0012 &
  \includegraphics[height=19mm]{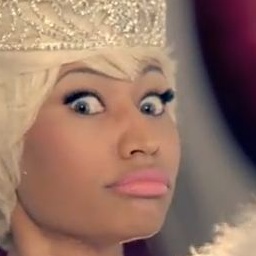} &
  \includegraphics[height=19mm]{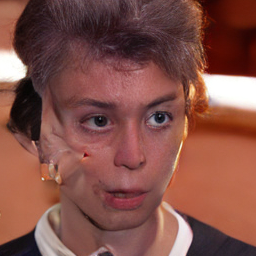} &
  \includegraphics[height=19mm]{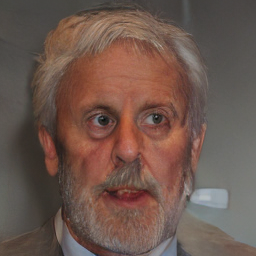} &
  \includegraphics[height=19mm]{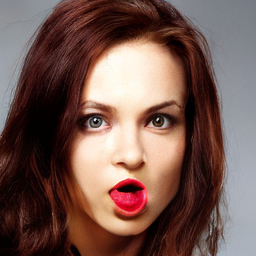} &
  \includegraphics[height=19mm]{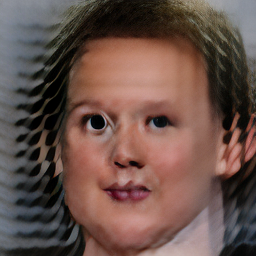} &
  \includegraphics[height=19mm]{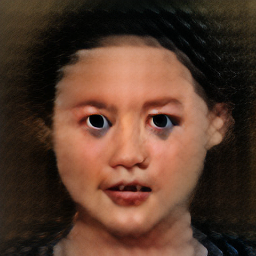} &
  \includegraphics[height=19mm]{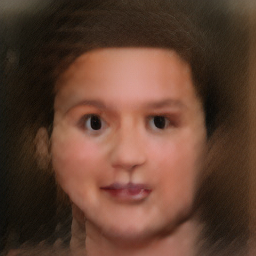} &
  \includegraphics[height=19mm]{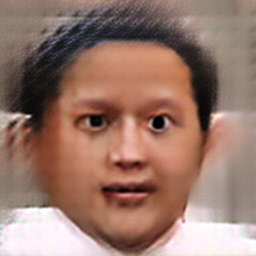} \\[2pt]
  b2\_0013 &
  \includegraphics[height=19mm]{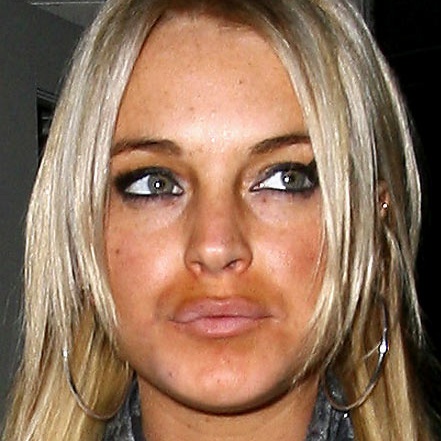} &
  \includegraphics[height=19mm]{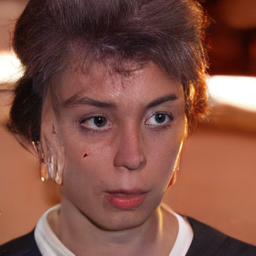} &
  \includegraphics[height=19mm]{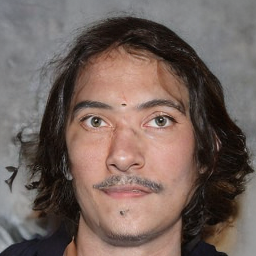} &
  \includegraphics[height=19mm]{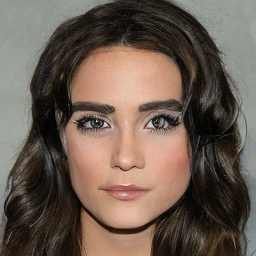} &
  \includegraphics[height=19mm]{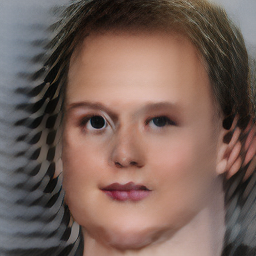} &
  \includegraphics[height=19mm]{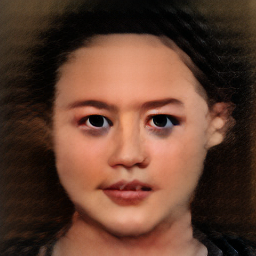} &
  \includegraphics[height=19mm]{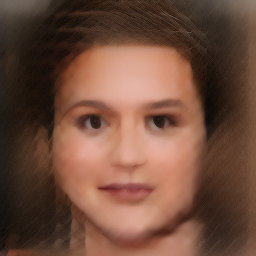} &
  \includegraphics[height=19mm]{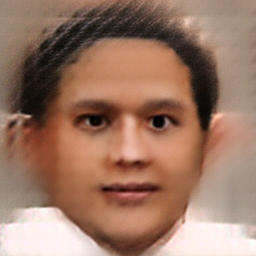} \\[2pt]
  b2\_0014 &
  \includegraphics[height=19mm]{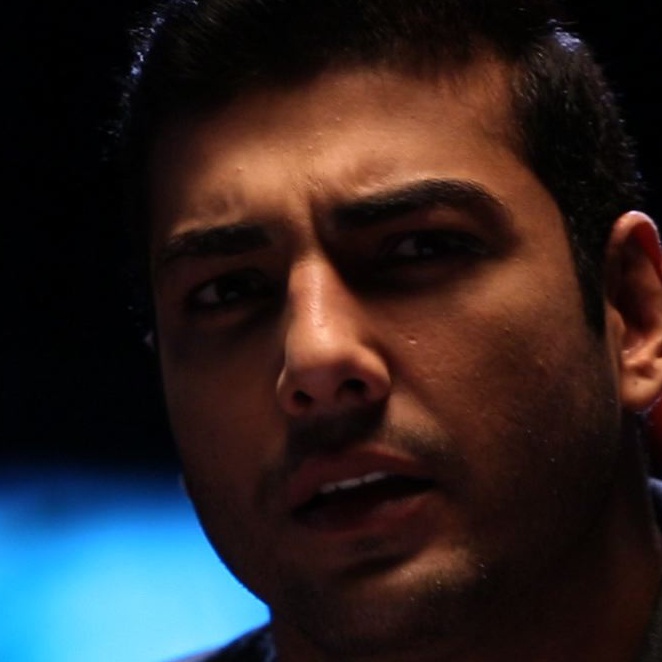} &
  \includegraphics[height=19mm]{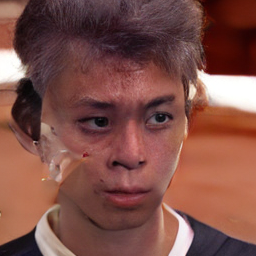} &
  \includegraphics[height=19mm]{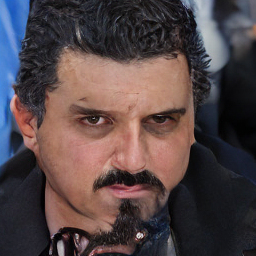} &
  \includegraphics[height=19mm]{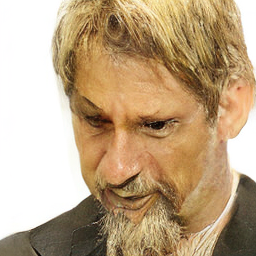} &
  \includegraphics[height=19mm]{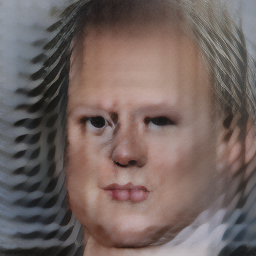} &
  \includegraphics[height=19mm]{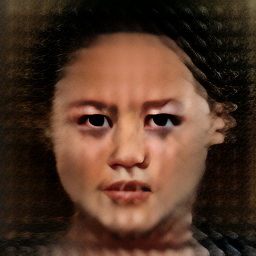} &
  \includegraphics[height=19mm]{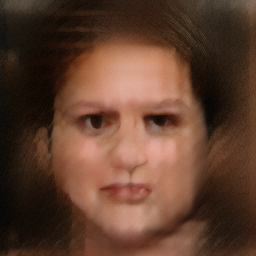} &
  \includegraphics[height=19mm]{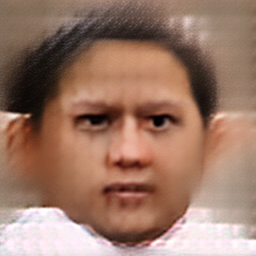} \\[2pt]
  b2\_0015 &
  \includegraphics[height=19mm]{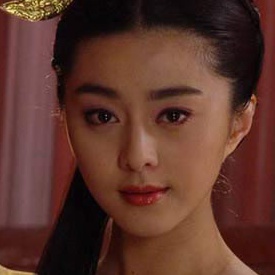} &
  \includegraphics[height=19mm]{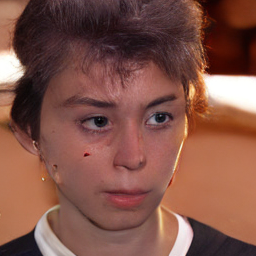} &
  \includegraphics[height=19mm]{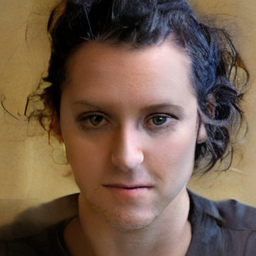} &
  \includegraphics[height=19mm]{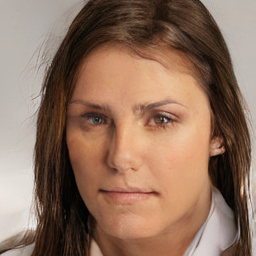} &
  \includegraphics[height=19mm]{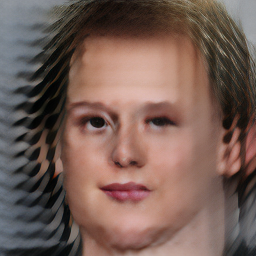} &
  \includegraphics[height=19mm]{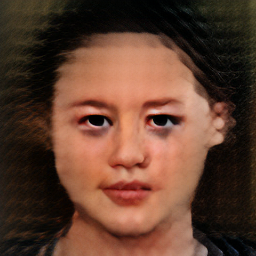} &
  \includegraphics[height=19mm]{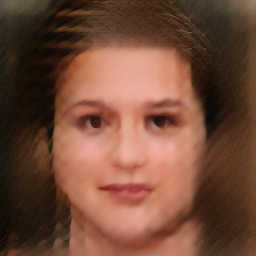} &
  \includegraphics[height=19mm]{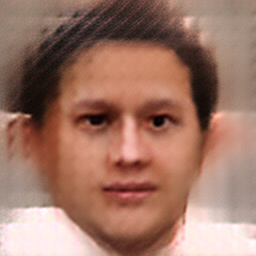} \\[2pt]
  b2\_0016 &
  \includegraphics[height=19mm]{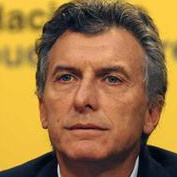} &
  \includegraphics[height=19mm]{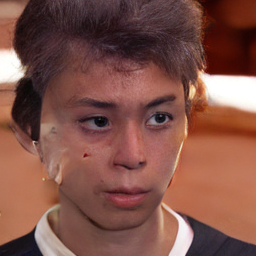} &
  \includegraphics[height=19mm]{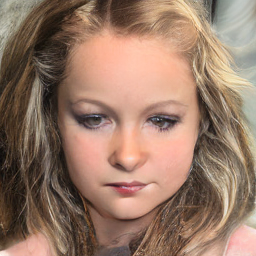} &
  \includegraphics[height=19mm]{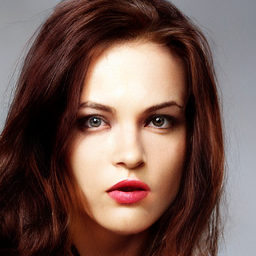} &
  \includegraphics[height=19mm]{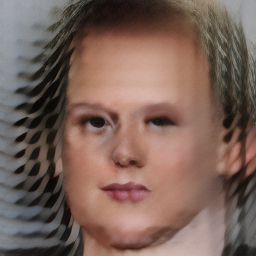} &
  \includegraphics[height=19mm]{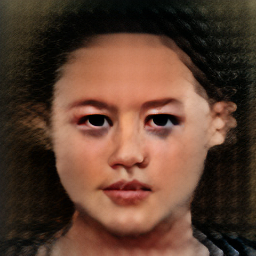} &
  \includegraphics[height=19mm]{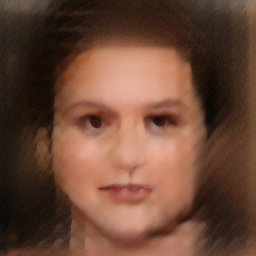} &
  \includegraphics[height=19mm]{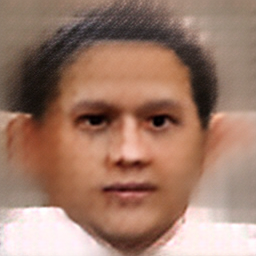} \\
  \end{tabular}}

  \caption{\textbf{Additional sources.} Rows are sources b2\_0012--b2\_0016. Columns: source image, random sampling, two SAFA outputs, PCGrad, CAGrad, FAMO, and AlignProp, all from MeanFlow B/2. Random and optimization baselines use noise index 0 throughout; SAFA shows its top two outputs per source.}
  \label{fig:supp:gallery_c}
\end{figure*}

\renewcommand{\thefigure}{S\arabic{figure}}
\setcounter{figure}{6}

\section{Main-Paper Results and Reproducibility Details}
\label{sup:sec:repro}

\subsection{Default selection at an eight-output budget}
\label{sup:sec:repro-default}

Table~\ref{sup:tab:default-summary} reproduces the default results from main-paper Tables~1 and~4, including the two held-out encoders and their mean. \safa{} ranks predicted noise scores and generates only eight images, whereas BoN generates and evaluates all 128 candidates before returning eight, giving a generation ratio of $8/128=6.25\%$. Main-paper \safa{} achieves $\bar{s}=0.682$, versus $0.677$ for unpenalized BoN and $0.672$ for penalized BoN.

\begin{table*}[t]
\centering
\small
\setlength{\tabcolsep}{5pt}
\renewcommand{\arraystretch}{1.10}

\caption{Default results from main-paper Tables~1 and~4, with $\bar{s}$ computed as the mean of the reported held-out encoder similarities. Each method returns eight outputs per source for all 512 test sources. $q$ is a percentage, and $U_8$ is measured in images per source.}
\label{sup:tab:default-summary}
\begin{tabular}{@{}lcrrrrrr@{}}
\toprule
Method & $w$ & $s_0$ & $s_{E_A}$ & $s_{E_B}$ & $\bar{s}$ & $q$ (\%) & $U_8$ \\
\midrule
BoN & 0 & 0.908 & 0.689 & 0.665 & 0.677 & 65.9 & 4.97 \\
BoN & 0.2 & 0.892 & 0.685 & 0.659 & 0.672 & 95.9 & 6.79 \\
\safa{} & 0.2 & 0.885 & 0.694 & 0.670 & 0.682 & 87.4 & 5.85 \\
\bottomrule
\end{tabular}
\end{table*}

\subsection{Expression supervision and classifier controls}
\label{sup:sec:classifier-controls}

The upstream expression encoder $E_0$ is a ResNet-50 trained with eight-class expression supervision and produces a 512-dimensional representation. It remains frozen during generator conditioning, response computation, and selector training.

The supplementary experiments in Tables~\ref{sup:tab:raf-transfer} and~\ref{sup:tab:fer-class} use separate classifier runs under fixed per-source augmentation quotas to analyze class-specific recognition and assess cross-dataset transfer, sharing the same real data split as main-paper Table~2. Classifiers use 1,200 AffectNet-S training images (150 Neutral, 50 Happiness, 50 Sadness, 50 Surprise, 300 Fear, 150 Disgust, 300 Anger, and 150 Contempt across standard class IDs 0--7), 800 development images, and the 3,969-image AffectNet validation set; the Real-only and +\safa{} classifiers are additionally evaluated on the 3,071-image RAF-DB test set. Augmented runs retain all 1,200 real images and add 5,071 generated training images (6,271 training images per arm) under a fixed, shared per-source quota $n_i \in \{0,\dots,8\}$ across the 1,200 training sources (seven sources receive $n_i=0$ while retaining their real image); generated images inherit their source expression label. Because the per-source quotas are shared, all augmented arms add identical per-class counts: 690 Neutral, 291 Happiness, 193 Sadness, 210 Surprise, 885 Fear, 634 Disgust, 1,271 Anger, and 897 Contempt images, totaling 5,071.

ResNet-18 is initialized from ImageNet \texttt{resnet18.a1\_in1k} and ConvNeXt-T from \texttt{convnext\_tiny.in12k\_ft\_in1k}, with eight-class classification heads. Training runs for 2,200 updates with batch sizes of 128 and 96, respectively, using AdamW (learning rate $3\times10^{-4}$, weight decay $10^{-4}$), class-weighted cross-entropy, and FP32 precision with TF32 disabled. ResNet-18 uses a cosine learning-rate decay, whereas ConvNeXt-T uses a stage-wise learning-rate decay factor of 0.7, a 200-update warmup, and cosine decay. Training images are randomly cropped using a shared width-and-height scale sampled uniformly from 0.7 to 1.0, horizontally flipped with probability 0.5, resized to $224\times224$, and normalized using ImageNet statistics. The loss weight for class $c$ is $n/(8n_c)$, where $n$ and $n_c$ count all training images and class-$c$ images in the corresponding training arm, including generated images when used. Development macro-F1 is evaluated every 200 updates, selecting the checkpoint at the earliest maximum. Classifier seeds 0--4 are independent of selector seeds, and RAF-DB evaluation reuses the AffectNet-selected checkpoints without selection on RAF-DB.

For the comparison in Table~\ref{sup:tab:fer-class}, +\safa{} ranks the $K=128$ noise dictionary using predicted scores from its selector trained with $w=0.2$, whereas the offline BoN reference generates and ranks all $K=128$ candidates per training source. To fill the per-source quota $n_i$, each method first searches its top 8 ranked candidates for a subset of size $n_i$ whose images satisfy the quality gate $Q(y)=1$ and pairwise LPIPS $\ge \delta$ ($\delta=0.127$); if fewer than $n_i$ feasible outputs are available among the top 8, the ranked prefix is expanded one candidate at a time up to at most rank 32, stopping as soon as $n_i$ feasible outputs are obtained. Under this quality- and distinctness-filtered quota filling, ranking the 128-candidate pool by measured $s_0$ ($w=0$) or by the quality-adjusted score ($w=0.2$) selects the exact same 5,071 training images and yields identical recognition results (reported simply as BoN in Table~\ref{sup:tab:fer-class}). For +\safa{}, only two training sources required inspection beyond rank 8, and both filled their quotas at rank 9.

\end{document}